\documentclass[final]{ws-ijprai}
\usepackage[bookmarks=false,pdfstartview={FitH}]{hyperref}
\usepackage{url}
\usepackage{graphicx, color}
\usepackage{amsmath, amssymb}
\usepackage{multirow}
\usepackage{xtab}
\usepackage{subfigure}

\newcommand{\bX}{{\mathcal X}}
\newcommand{\bx}{{\bf x}}
\newcommand{\bc}{{\bf c}}
\newcommand{\ul}{\underline}

\begin{document}

\markboth{M.\ Emre Celebi and Hassan A.\ Kingravi}
{Deterministic Initialization of the K-Means Algorithm}

%
\catchline{}{}{}{}{}
%

\title{Deterministic Initialization of the K-Means Algorithm Using Hierarchical Clustering}

\author{M.\ Emre Celebi}
\address{Department of Computer Science\\Louisiana State University, Shreveport, LA, USA\\
\email{ecelebi@lsus.edu}}

\author{Hassan A.\ Kingravi}
\address{School of Electrical and Computer Engineering\\Georgia Institute of Technology, Atlanta, GA, USA\\
\email{kingravi@gatech.edu}}

\maketitle

\begin{abstract}
K-means is undoubtedly the most widely used partitional clustering algorithm. Unfortunately, due to its gradient descent nature, this algorithm is highly sensitive to the initial placement of the cluster centers. Numerous initialization methods have been proposed to address this problem. Many of these methods, however, have superlinear complexity in the number of data points, making them impractical for large data sets. On the other hand, linear methods are often random and/or order-sensitive, which renders their results unrepeatable. Recently, Su and Dy proposed two highly successful hierarchical initialization methods named Var-Part and PCA-Part that are not only linear, but also deterministic (non-random) and order-invariant. In this paper, we propose a discriminant analysis based approach that addresses a common deficiency of these two methods. Experiments on a large and diverse collection of data sets from the UCI Machine Learning Repository demonstrate that Var-Part and PCA-Part are highly competitive with one of the best random initialization methods to date, i.e., k-means++, and that the proposed approach significantly improves the performance of both hierarchical methods.
\end{abstract}


\section{Introduction}
\label{sec_intro}

Clustering, the unsupervised classification of patterns into groups, is one of the most important tasks in exploratory  data analysis \cite{Jain99}. Primary goals of clustering include gaining insight into data (detecting anomalies, identifying salient features, etc.), classifying data, and compressing data. Clustering has a long and rich history in a variety of scientific disciplines including anthropology, biology, medicine, psychology, statistics, mathematics, engineering, and computer science. As a result, numerous clustering algorithms have been proposed since the early 1950s \cite{Jain10}.
\par
Clustering algorithms can be broadly classified into two groups: hierarchical and partitional \cite{Jain10}. Hierarchical algorithms recursively find nested clusters either in a top-down (divisive) or bottom-up (agglomerative) fashion. In contrast, partitional  algorithms find all the clusters simultaneously as a partition of the data and do not impose a hierarchical structure. Most hierarchical algorithms have quadratic or higher complexity in the number of data points \cite{Jain99} and therefore are not suitable for large data sets, whereas partitional algorithms often have lower complexity.
\par
Given a data set $\bX = \{ \bx_1, \bx_2, \dotsc, \bx_N \}$ in $\mathbb{R}^D$, i.e., $N$ points (vectors) each with $D$ attributes (components), hard partitional algorithms divide $\bX$ into $K$ exhaustive and mutually exclusive clusters $\mathcal{P} = \{ P_1, P_2, \dotsc, P_K \},$ $\;\; \bigcup\nolimits_{i = 1}^K {P_i = \bX},$ $\;\; P_i \cap P_j = \emptyset$ for $1 \leq i \neq j \leq K$. These algorithms usually generate clusters by optimizing a criterion function. The most intuitive and frequently used criterion function is the Sum of Squared Error (SSE) given by:
\begin{equation}
\label{eq_sse}
 \mbox{SSE} = \sum\limits_{i = 1}^K {\sum\limits_{\bx_j \in P_i} {\left\| {\bx_j - \bc_i } \right\|_2^2 } }
\end{equation}
where $\| . \|_2$ denotes the Euclidean ($\mathcal{L}_2$) norm and
$\bc_i =
{1 \mathord{\left/ {\vphantom {1 {\left| P_i \right|}}} \right.
 \kern-\nulldelimiterspace} {\left| P_i \right|}}\sum\nolimits_{\bx_j \in P_i} {\bx_j}$ is the centroid of cluster $P_i$ whose cardinality is $\left| P_i \right|$. The optimization of \eqref{eq_sse} is often referred to as the minimum SSE clustering (MSSC) problem.

The number of ways in which a set of $N$ objects can be partitioned into $K$ non-empty groups is given by Stirling numbers of the second kind:
\begin{equation}
\label{eq_num_parts}
\mathcal{S}(N,K) =
\frac{1}
{{K!}}\sum\limits_{i = 0}^K {( - 1)^{K - i} \left( \begin{gathered}
  K \hfill \\
  i \hfill \\
\end{gathered} \right)} i^N
\end{equation}
which can be approximated by $K^N/K!$ It can be seen that a complete enumeration of all possible clusterings to determine the global minimum of \eqref{eq_sse} is clearly computationally prohibitive except for very small data sets \cite{Kaufman90}. In fact, this non-convex optimization problem is proven to be NP-hard even for $K = 2$ \cite{Aloise09} or $D = 2$ \cite{Mahajan12}. Consequently, various heuristics have been developed to provide approximate solutions to this problem \cite{Tarsitano03}. Among these heuristics, Lloyd's algorithm \cite{Lloyd82}, often referred to as the (batch) k-means algorithm, is the simplest and most commonly used one. This algorithm starts with $K$ arbitrary centers, typically chosen uniformly at random from the data points. Each point is assigned to the nearest center and then each center is recalculated as the mean of all points assigned to it. These two steps are repeated until a predefined termination criterion is met.
\par
The k-means algorithm is undoubtedly the most widely used partitional clustering algorithm \cite{Jain10}. Its popularity can be attributed to several reasons. First, it is conceptually simple and easy to implement. Virtually every data mining software includes an implementation of it. Second, it is versatile, i.e., almost every aspect of the algorithm (initialization, distance function, termination criterion, etc.) can be modified. This is evidenced by hundreds of publications over the last fifty years that extend k-means in various ways. Third, it has a time complexity that is linear in $N$, $D$, and $K$ (in general, $D \ll N$ and $K \ll N$). For this reason, it can be used to initialize more expensive clustering algorithms such as expectation maximization \cite{Bradley98}, DBSCAN \cite{Dash01}, and spectral clustering \cite{Chen11}. Furthermore, numerous sequential \cite{Kanungo02,Hamerly10} and parallel \cite{Chen10} acceleration techniques are available in the literature. Fourth, it has a storage complexity that is linear in $N$, $D$, and $K$. In addition, there exist disk-based variants that do not require all points to be stored in memory \cite{Ordonez04}. Fifth, it is guaranteed to converge \cite{Selim84} at a quadratic rate \cite{Bottou95}. Finally, it is invariant to data ordering, i.e., random shufflings of the data points.
\par
On the other hand, k-means has several significant disadvantages. First, it requires the number of clusters, $K$, to be specified in advance. The value of this parameter can be determined automatically by means of various internal/relative cluster validity measures \cite{Vendramin10}. Second, it can only detect compact, hyperspherical clusters that are well separated. This can be alleviated by using a more general distance function such as the Mahalanobis distance, which permits the detection of hyperellipsoidal clusters \cite{Mao96}. Third, due its utilization of the squared Euclidean distance, it is sensitive to noise and outlier points since even a few such points can significantly influence the means of their respective clusters. This can be addressed by outlier pruning \cite{Zhang03} or using a more robust distance function such as City-block ($\mathcal{L}_1$) distance. Fourth, due to its gradient descent nature, it often converges to a local minimum of the criterion function \cite{Selim84}. For the same reason, it is highly sensitive to the selection of the initial centers \cite{Celebi13a}. Adverse effects of improper initialization include empty clusters, slower convergence, and a higher chance of getting stuck in bad local minima \cite{Celebi11}. Fortunately, except for the first two, these drawbacks can be remedied by using an adaptive initialization method (IM).
\par
A large number of IMs have been proposed in the literature \cite{Pena99,He04,Celebi11,Celebi13a}. Unfortunately, many of these have superlinear complexity in $N$ \cite{Lance67,Astrahan70,Hartigan79,Kaufman90,Likas03,AlDaoud05,Redmond07,AlHasan09,Cao09,Kang09}, which makes them impractical for large data sets (note that k-means itself has linear complexity). In contrast, linear IMs are often random and/or order-sensitive \cite{Forgy65,Jancey66,MacQueen67,Ball67,Tou74,Spath77,Bradley98,Arthur07}, which renders their results unrepeatable. Su and Dy proposed two divisive hierarchical initialization methods named Var-Part and PCA-Part that are not only linear, but also deterministic and order-invariant \cite{Su07}. In this study, we propose a simple modification to these methods that improves their performance significantly.
\par
The rest of the paper is organized as follows. Section \ref{sec_init_methods} presents a brief overview of some of the most popular linear, order-invariant k-means IMs and the proposed modification to Var-Part and PCA-Part. Section \ref{sec_exp} presents the experimental results, while Section \ref{sec_disc} analyzes these results. Finally, Section \ref{sec_conc} gives the conclusions.

\section{Linear, Order-Invariant Initialization Methods for K-Means}
\label{sec_init_methods}

\subsection{Overview of the Existing Methods}
Forgy's method \cite{Forgy65} assigns each point to one of the $K$ clusters uniformly at random. The centers are then given by the centroids of these initial clusters. This method has no theoretical basis, as such random clusters have no internal homogeneity \cite{Anderberg73}.

MacQueen \cite{MacQueen67} proposed two different methods. The first one, which is the default option in the Quick Cluster procedure of IBM SPSS Statistics \cite{Norusis11}, takes the first $K$ points in $\bX$ as the centers. An obvious drawback of this method is its sensitivity to data ordering. The second method chooses the centers randomly from the data points. The rationale behind this method is that random selection is likely to pick points from dense regions, i.e., points that are good candidates to be centers. However, there is no mechanism to avoid choosing outliers or points that are too close to each other \cite{Anderberg73}. Multiple runs of this method is the standard way of initializing k-means \cite{Bradley98}. It should be noted that this second method is often mistakenly attributed to Forgy \cite{Forgy65}.

The maximin method \cite{Gonzalez85} chooses the first center $\bc_1$ arbitrarily and the $i$-th ($i \in \{2, 3, \dotsc, K\}$) center $\bc_i$ is chosen to be the point that has the greatest minimum-distance to the previously selected centers, i.e., $\bc_1, \bc_2, \dotsc, \bc_{i-1}$. This method was originally developed as a $2$-approximation to the $K$-center clustering problem\footnote{Given a set of $N$ points in a metric space, the goal of $K$-center clustering is to find $K$ representative points (centers) such that the maximum distance of a point to a center is minimized. Given a minimization problem, a \emph{$2$-approximation} algorithm is one that finds a solution whose cost is at most twice the cost of the optimal solution.}.

The k-means++ method \cite{Arthur07} interpolates between MacQueen's second method and the maximin method. It chooses the first center randomly and the $i$-th ($i \in \{ 2, 3, \dotsc, K \}$) center is chosen to be $\bx' \in \bX$ with a probability of $\frac{{md\left( {\bx'} \right)^2 }}{{\sum\nolimits_{j = 1}^N {md(\bx_j)^2 }}}$, where $md(\bx)$ denotes the minimum-distance from a point $\bx$ to the previously selected centers. This method yields an $\Theta(\log{K})$ approximation to the MSSC problem.

The PCA-Part method \cite{Su07} uses a divisive hierarchical approach based on PCA (Principal Component Analysis) \cite{Hotelling36}. Starting from an initial cluster that contains the entire data set, the method iteratively selects the cluster with the greatest SSE and divides it into two subclusters using a hyperplane that passes through the cluster centroid and is orthogonal to the principal eigenvector of the cluster covariance matrix. This procedure is repeated until $K$ clusters are obtained. The centers are then given by the centroids of these clusters. The Var-Part method \cite{Su07} is an approximation to PCA-Part, where the covariance matrix of the cluster to be split is assumed to be diagonal. In this case, the splitting hyperplane is orthogonal to the coordinate axis with the greatest variance.

\begin{figure}[!ht]
\centering
 \subfigure[Input data set]{\includegraphics[width=0.48\columnwidth]{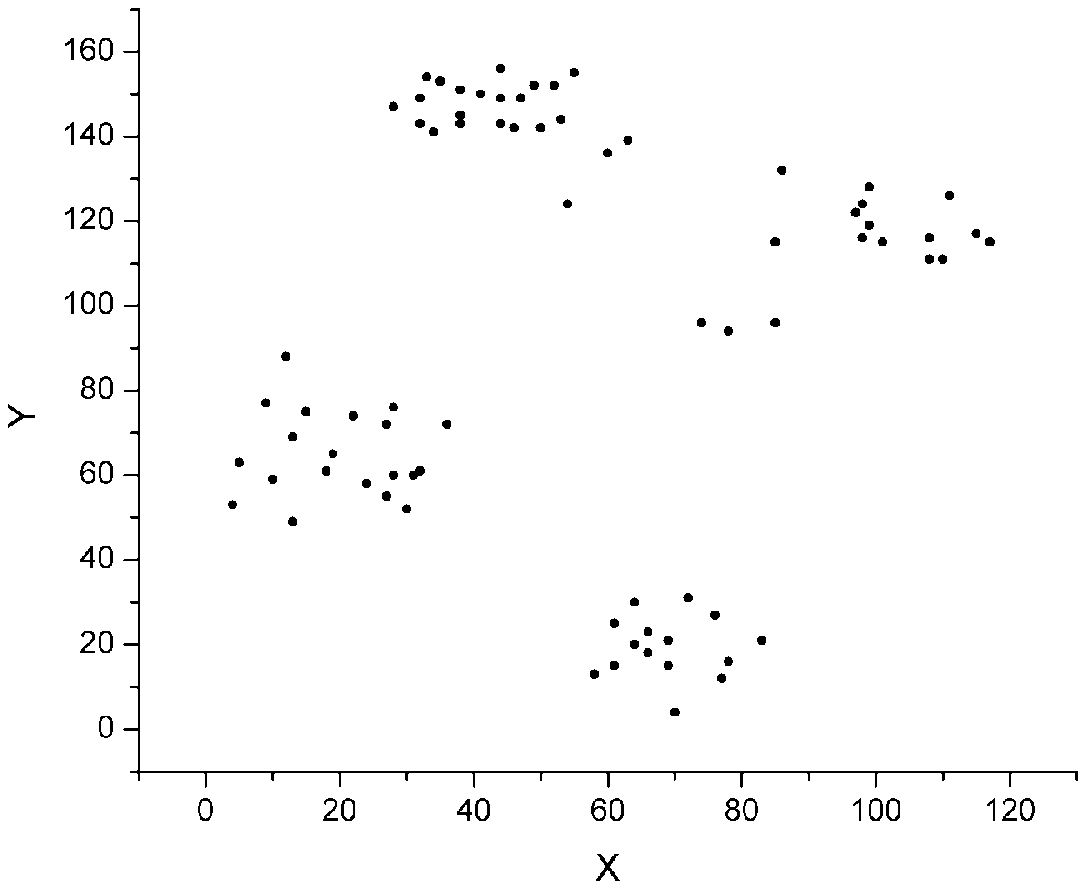}}
 \subfigure[Iteration 1]{\includegraphics[width=0.48\columnwidth]{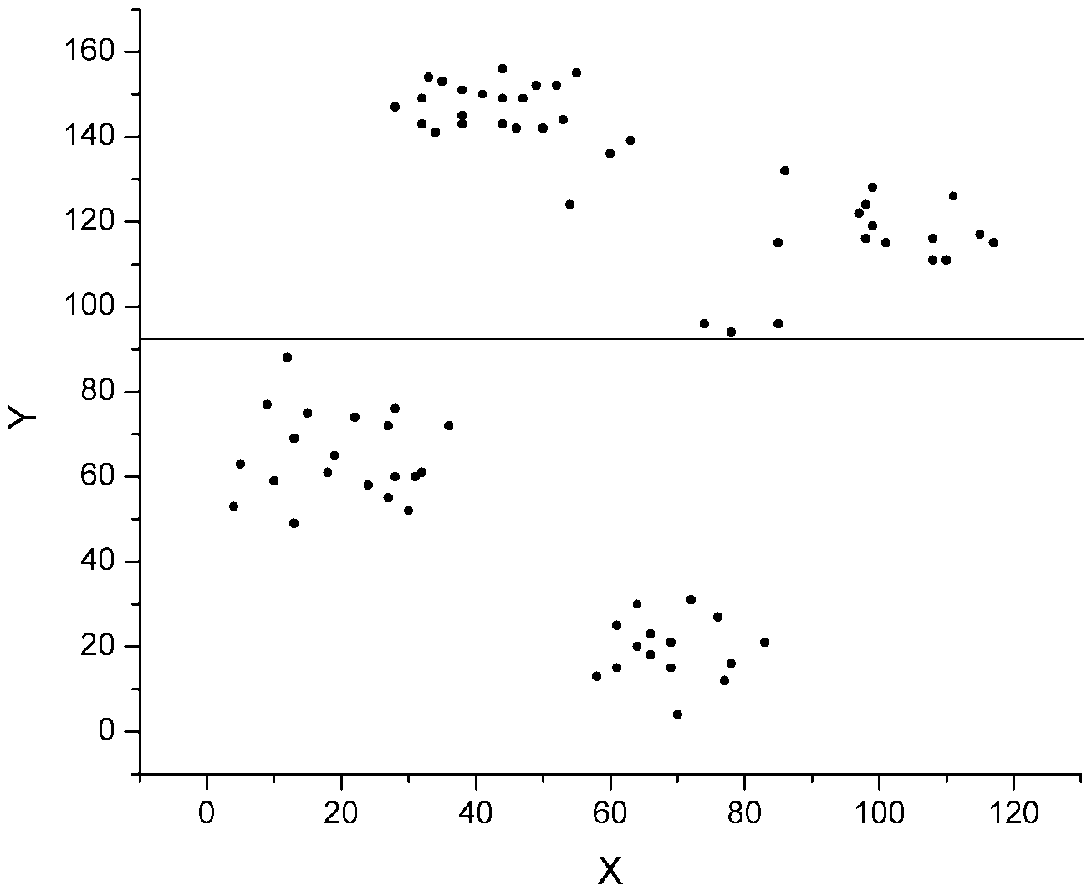}}
 \\
 \subfigure[Iteration 2]{\includegraphics[width=0.48\columnwidth]{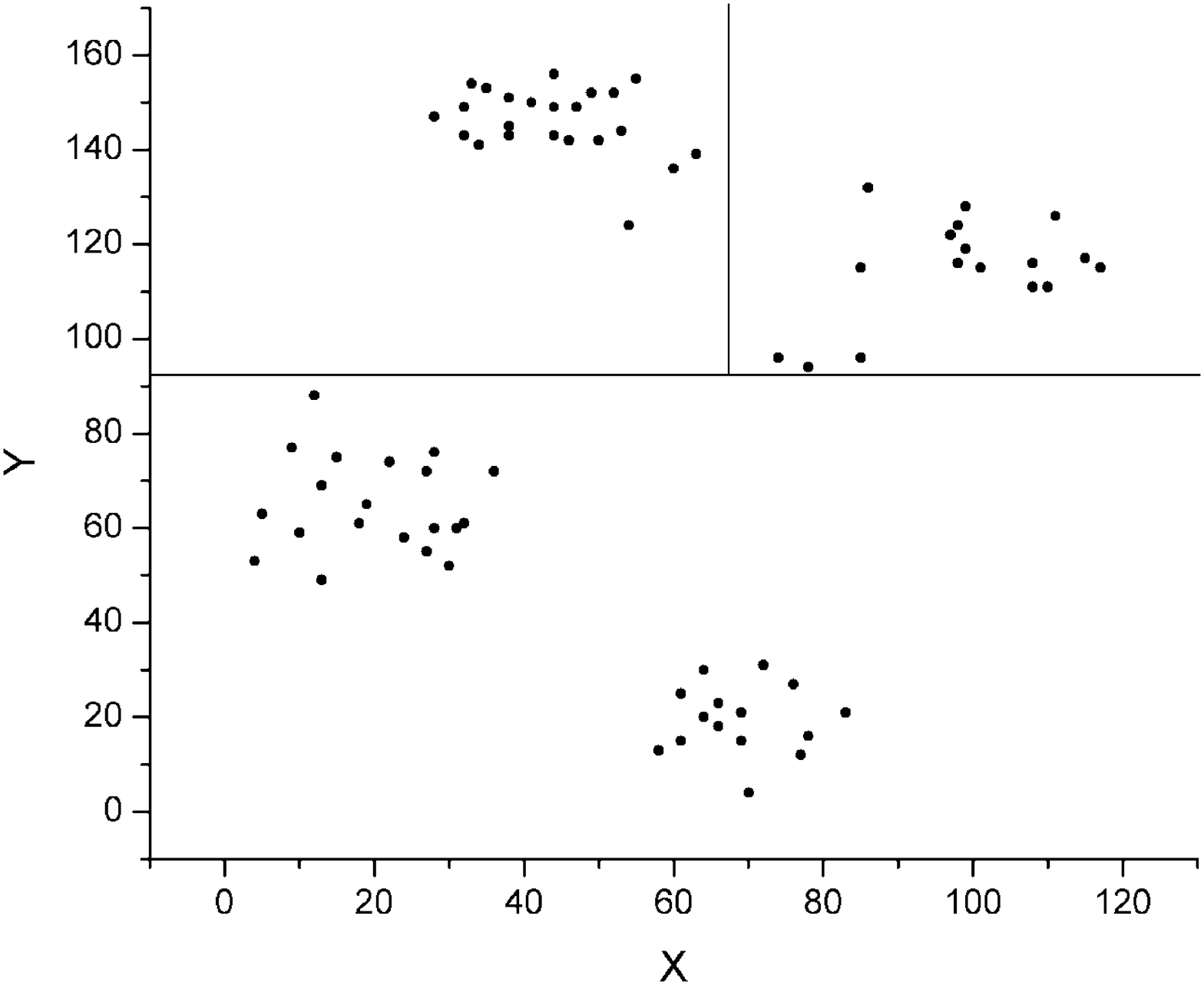}}
 \subfigure[Iteration 3]{\label{part_d} \includegraphics[width=0.48\columnwidth]{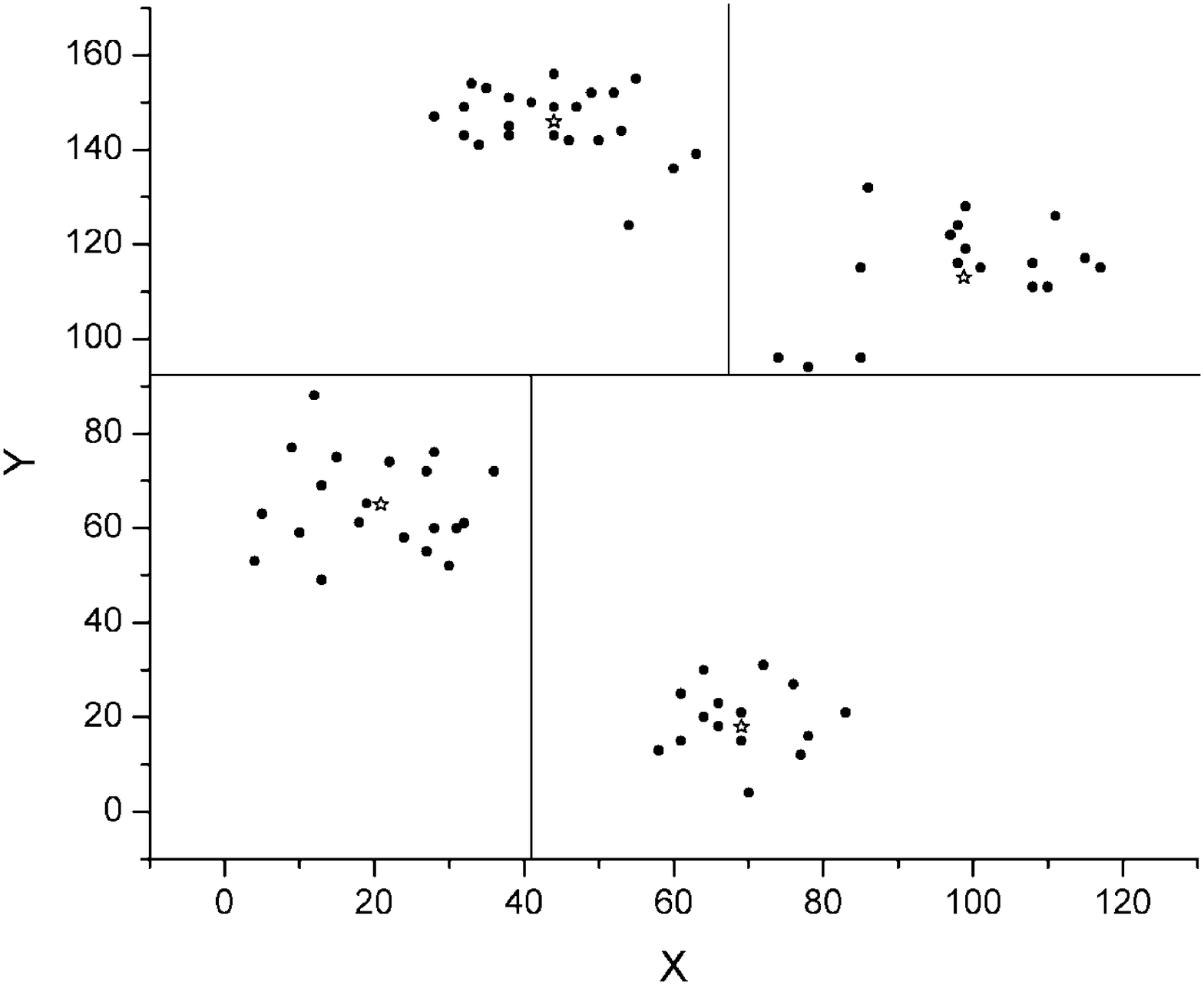}}
 \caption{Illustration of Var-Part on the Ruspini data set}
 \label{fig_varpart_ruspini}
\end{figure}

Figure \ref{fig_varpart_ruspini} illustrates the Var-Part procedure on a toy data set with four natural clusters \cite{Ruspini70}. In iteration $1$, the initial cluster that contains the entire data set is split into two subclusters along the Y axis using a line (one-dimensional hyperplane) that passes through the mean point ($92.026667$). Between the resulting two clusters, the one above the line has a greater SSE. In iteration $2$, this cluster is therefore split along the X axis at the mean point ($66.975000$). In the final iteration, the cluster with the greatest SSE, i.e., the bottom cluster, is split along the X axis at the mean point ($41.057143$). In Figure \ref{part_d}, the centroids of the final four clusters are denoted by stars.

\subsection{Proposed Modification to Var-Part and PCA-Part}
\label{sec_mod}

Su and Dy \cite{Su07} demonstrated that, besides being computationally efficient, Var-Part and PCA-Part perform very well on a variety of data sets. Recall that in each iteration these methods select the cluster with the greatest SSE and then project the $D$-dimensional points in this cluster on a partitioning axis. The difference between the two methods is the choice of this axis. In Var-Part, the partitioning axis is the coordinate axis with the greatest variance, whereas in PCA-Part it is the major axis. After the projection operation, both methods use the \emph{mean} point on the partitioning axis as a `threshold' to divide the points between two clusters. In other words, each point is assigned to one of the two subclusters depending on which side of the \emph{mean} point its projection falls to. It should be noted that the choice of this threshold is primarily motivated by computational convenience. Here, we propose a better alternative based on discriminant analysis.

The projections of the points on the partitioning axis can be viewed as a discrete probability distribution, which can be conveniently represented by a histogram. The problem of dividing a histogram into two partitions is a well studied one in the field of image processing. A plethora of histogram partitioning, a.k.a. thresholding, methods has been proposed in the literature with the early ones dating back to the 1960s \cite{Sezgin04}. Among these, Otsu's method \cite{Otsu79} has become the method of choice as confirmed by numerous comparative studies \cite{Sahoo88,Lee90,Trier95,Sezgin04,Medina05,Celebi13b}.

Given an image represented by $L$ gray levels $\left\{ 0, 1, \dotsc, L - 1\right\}$, a thresholding method partitions the image pixels into two classes $\mathcal{C}_0 = \left\{ 0, 1, \dotsc, t \right\}$ and $\mathcal{C}_1 = \{ t + 1, t + 2,$ $\dotsc, L - 1 \}$ (object and background, or vice versa) at gray level $t$. In other words, pixels with gray levels less than or equal to the threshold $t$ are assigned to $\mathcal{C}_0$, whereas the remaining pixels are assigned to $\mathcal{C}_1$.

Let $n_i$ be the number of pixels with gray level $i$. The total number of pixels in the image is then given by $n = \sum\nolimits_{i = 0}^{L - 1}{n_i}$. The normalized gray level histogram of the image can be regarded as a probability mass function:
\begin{equation*}
p_i = \frac{n_i}{n},\quad p_i \ge 0,\quad \sum\limits_{i = 0}^{L - 1} {p_i}  = 1
\end{equation*}

Let $p_0(t) = \sum\nolimits_{i = 0}^t {p_i}$ and $p_1(t) = 1 - p_0(t)$ denote the probabilities of $\mathcal{C}_0$ and $\mathcal{C}_1$, respectively. The means of the respective classes are then given by:
\begin{equation*}
\begin{array}{l}
 \mu _0 (t) = {{\mu (t)} \mathord{\left/
 {\vphantom {{\mu (t)} {p_0 (t)}}} \right.
 \kern-\nulldelimiterspace} {p_0 (t)}} \\
 \mu _1 (t) = {{\left( {\mu _T  - \mu (t)} \right)} \mathord{\left/
 {\vphantom {{\left( {\mu _T  - \mu (t)} \right)} {p_1 (t)}}} \right.
 \kern-\nulldelimiterspace} {p_1 (t)}} \\
 \end{array}
\end{equation*}
where $\mu(t) = \sum\nolimits_{i = 0}^t {ip_i}$ and $\mu_T  = \mu(L - 1)$ denote the first moment of the histogram up to gray level $t$ and mean gray level of the image, respectively.

Otsu's method adopts between-class variance, i.e., $\sigma_B^2(t) = p_0(t) p_1(t) \left[ {\mu_0(t) - \mu_1(t)} \right]^2$, from the discriminant analysis literature as its objective function and determines the optimal threshold $t^*$ as the gray level that maximizes $\sigma_B^2(t)$, i.e., $t^* = \mathop {{\mathop{\rm argmax}\nolimits} }\limits_{t \in \left\{0,1,\dotsc,L - 1 \right\}} \sigma_B^2(t)$. Between-class variance can be viewed as a measure of class separability or histogram bimodality. It can be efficiently calculated using:
\begin{equation*}
\sigma_B^2(t) = \frac{{\left[ {\mu_T p_0(t) - \mu(t)} \right]^2 }}{{p_0(t) p_1(t)}}
\end{equation*}

It should be noted that the efficiency of Otsu's method can be attributed to the fact that it operates on histogrammed pixel gray values, which are non-negative integers. Var-Part and PCA-Part, on the other hand, operate on the projections of the points on the partitioning axis, which are often fractional. This problem can be circumvented by linearly scaling the projection values to the limits of the histogram, i.e., $0$ and $L-1$. Let $y_i$ be the projection of a point $\bx_i$ on the partitioning axis. $y_i$ can be mapped to histogram bin $b$ given by:
\begin{equation*}
b = \left\lfloor {\frac{{L\left( {y_i - \mathop {\min }\limits_j y_j } \right)}}{{\mathop {\max }\limits_j y_j - \mathop {\min}\limits_j y_j }}} \right\rfloor
\end{equation*}
where $\lfloor z \rfloor$ is the floor function which returns the largest integer less than or equal to $z$.

The computational complexities of histogram construction and Otsu's method are $\mathcal{O}(N_i)$ ($N_i$: number of points in the cluster) and $\mathcal{O}(L)$, respectively. $L$ is constant in our experiments and therefore the proposed modification does not alter the linear time complexity of Var-Part and PCA-Part.

\begin{figure}[!ht]
\centering
\includegraphics[width=0.8\columnwidth]{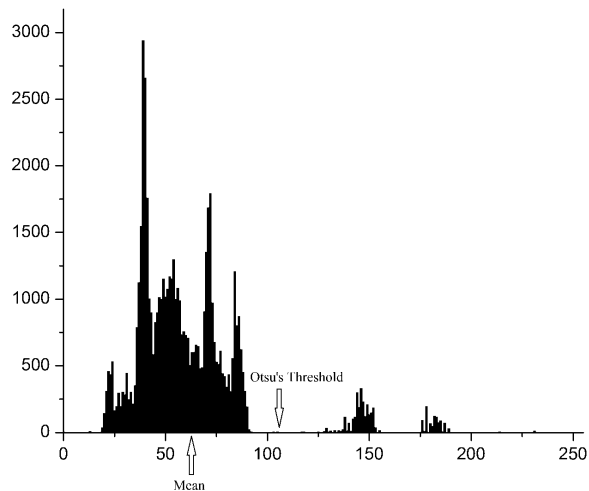}
\caption{\label{fig_histo} Comparison of mean point and Otsu's thresholds}
\end{figure}

Figure \ref{fig_histo} shows a histogram where using the mean point as a threshold leads to poor results. This histogram is constructed during the first iteration of PCA-Part from the projections of the points in the Shuttle data set (see Table \ref{tab_data_set}). As marked on the figure, the mean point of this histogram is $61$, whereas Otsu's method gives a threshold of $105$. The SSE of the initial cluster is $1,836$. When the mean point of the histogram is used a threshold, the resulting two subclusters have SSE's of $408$ and $809$. This means that splitting the initial cluster with a hyperplane orthogonal to the principal eigenvector of the cluster covariance matrix at the mean point results in approximately $34$\% reduction in the SSE. On the other hand, when Otsu's threshold is used, the subclusters have SSE's of $943$ and $101$, which translates to about $43$\% reduction in the SSE. In the next section, we will demonstrate that using Otsu's threshold instead of the mean point often leads to significantly better initial clusterings on a variety of data sets.

\section{Experimental Results}
\label{sec_exp}

The experiments were performed on $24$ commonly used data sets from the UCI Machine Learning Repository \cite{Frank12}. Table \ref{tab_data_set} gives the data set descriptions. For each data set, the number of clusters ($K$) was set equal to the number of classes ($K'$), as commonly seen in the related literature \cite{He04,Arthur07,Su07,Redmond07,AlHasan09,Cao09,Kang09,Onoda12}.
\par
In clustering applications, normalization is a common preprocessing step that is necessary to prevent attributes with large ranges from dominating the distance calculations and also to avoid numerical instabilities in the computations. Two commonly used normalization schemes are linear scaling to unit range (min-max normalization) and linear scaling to unit variance (z-score normalization). Several studies revealed that the former scheme is preferable to the latter since the latter is likely to eliminate valuable between-cluster variation \cite{Milligan88,Su07}. As a result, we used min-max normalization to map the attributes of each data set to the $[0,1]$ interval.

\begin{table}[ht]
\centering
\caption{\label{tab_data_set} Data Set Descriptions ($N$: \# points, $D$: \# attributes, $K'$: \# classes)}
\begin{tabular}{llrrr}
\hline
ID & Data Set & $N$ & $D$ & $K'$\\
\hline
\hline
1 & Abalone & 4,177 & 7 & 28\\
2 & Breast Cancer Wisconsin (Original) & 683 & 9 & 2\\
3 & Breast Tissue & 106 & 9 & 6\\
4 & Ecoli & 336 & 7 & 8\\
5 & Glass Identification & 214 & 9 & 6\\
6 & Heart Disease & 297 & 13 & 5\\
7 & Ionosphere & 351 & 34 & 2\\
8 & Iris (Bezdek) & 150 & 4 & 3\\
9 & ISOLET & 7,797 & 617 & 26\\
10 & Landsat Satellite (Statlog) & 6,435 & 36 & 6\\
11 & Letter Recognition & 20,000 & 16 & 26\\
12 & MAGIC Gamma Telescope & 19,020 & 10 & 2\\
13 & Multiple Features (Fourier) & 2,000 & 76 & 10\\
14 & Musk (Clean2) & 6,598 & 166 & 2\\
15 & Optical Digits & 5,620 & 64 & 10\\
16 & Page Blocks Classification & 5,473 & 10 & 5\\
17 & Pima Indians Diabetes & 768 & 8 & 2\\
18 & Shuttle (Statlog) & 58,000 & 9 & 7\\
19 & Spambase & 4,601 & 57 & 2\\
20 & SPECTF Heart & 267 & 44 & 2\\
21 & Wall-Following Robot Navigation & 5,456 & 24 & 4\\
22 & Wine Quality & 6,497 & 11 & 7\\
23 & Wine & 178 & 13 & 3\\
24 & Yeast & 1,484 & 8 & 10\\
\hline
\end{tabular}
\end{table}

The performance of the IMs was quantified using two effectiveness (quality) and two efficiency (speed) criteria:

\begin{itemize}
\renewcommand{\labelitemi}{$\triangleright$}
 \item {\bf Initial SSE}: This is the SSE value calculated after the initialization phase, before the clustering phase. It gives us a measure of the effectiveness of an IM by itself.
 \item {\bf Final SSE}: This is the SSE value calculated after the clustering phase. It gives us a measure of the effectiveness of an IM when its output is refined by k-means. Note that this is the objective function of the k-means algorithm, i.e., \eqref{eq_sse}.
 \item {\bf Number of Iterations}: This is the number of iterations that k-means requires until reaching convergence when initialized by a particular IM. It is an efficiency measure independent of programming language, implementation style, compiler, and CPU architecture.
 \item {\bf CPU Time}: This is the total CPU time in milliseconds taken by the initialization and clustering phases.
\end{itemize}

All of the methods were implemented in the \texttt{C} language, compiled with the \texttt{gcc} v4.4.3 compiler, and executed on an Intel Xeon E5520 2.26GHz machine. Time measurements were performed using the \texttt{getrusage} function, which is capable of measuring CPU time to an accuracy of a microsecond. The MT19937 variant of the Mersenne Twister algorithm was used to generate high quality pseudorandom numbers \cite{Matsumoto98}.

The convergence of k-means was controlled by the disjunction of two criteria: the number of iterations reaches a maximum of $100$ or the relative improvement in SSE between two consecutive iterations drops below a threshold, i.e., $ {{\left( {\mbox{SSE}_{i - 1} - \mbox{SSE}_i} \right)} \mathord{\left/ {\vphantom {{\left( {\mbox{SSE}_{i - 1} - \mbox{SSE}_i} \right)} {\mbox{SSE}_i}}} \right. \kern-\nulldelimiterspace} {\mbox{SSE}_i}} \leq \varepsilon $, where $\mbox{SSE}_i$ denotes the SSE value at the end of the $i$-th ($i \in \{ 1, 2, \dotsc, 100 \}$) iteration. The convergence threshold was set to $\varepsilon = 10^{-6}$.

In this study, we focus on IMs that have time complexity linear in $N$. This is because k-means itself has linear complexity, which is perhaps the most important reason for its popularity. Therefore, an IM for k-means should not diminish this advantage of the algorithm. The proposed methods, named Otsu Var-Part (\texttt{OV}) and Otsu PCA-Part (\texttt{OP}), were compared to six popular, linear, order-invariant IMs: Forgy's method (\texttt{F}), MacQueen's second method (\texttt{M}), maximin (\texttt{X}), k-means++ (\texttt{K}), Var-Part (\texttt{V}), and PCA-Part (\texttt{P}). It should be noted that among these methods \texttt{F}, \texttt{M}, and \texttt{K} are random, whereas \texttt{X}\footnote{The first center is chosen as the centroid of the data set.}, \texttt{V}, \texttt{P}, \texttt{OV}, and \texttt{OP} are deterministic.

We first examine the influence of $L$ (number of histogram bins) on the performance of \texttt{OV} and \texttt{OP}. Tables \ref{tab_num_bins_var} and \ref{tab_num_bins_pca} show the initial and final SSE values obtained by respectively \texttt{OV} and \texttt{OP} for $L = 64, 128, 256, 512, 1024$ on four of the largest data sets (the best values are \ul{underlined}). It can be seen that the performances of both methods are relatively insensitive to the value of $L$. Therefore, in the subsequent experiments we report the results for $L = 256$.

\begin{table}[ht]
\centering
\caption{\label{tab_num_bins_var} Influence of $L$ on the effectiveness of Otsu Var-Part ($L$: \# histogram bins)}
\begin{tabular}{llrrrrr}
\hline
ID & Criterion & $L=64$ & $L=128$ & $L=256$ & $L=512$ & $L=1024$\\
\hline
\hline
9 & Initial SSE & \ul{143859} & 144651 & 144658 & 144637 & 144638\\
 & Final SSE & 118267 & 119127 & \ul{118033} & 118033 & 118034\\
\hline
10 & Initial SSE & 1987 & 1920 & \ul{1919} & 1920 & 1920\\
 & Final SSE & \ul{1742} & \ul{1742} & \ul{1742} & \ul{1742} & \ul{1742}\\
\hline
11 & Initial SSE & 3242 & \ul{3192} & 3231 & 3202 & 3202\\
 & Final SSE & 2742 & \ul{2734} & \ul{2734} & \ul{2734} & \ul{2734}\\
\hline
15 & Initial SSE & \ul{17448} & 17504 & 17504 & 17504 & 17504\\
 & Final SSE & \ul{14581} & \ul{14581} & \ul{14581} & \ul{14581} & \ul{14581}\\
\hline
\end{tabular}
\end{table}

\begin{table}[ht]
\centering
\caption{\label{tab_num_bins_pca} Influence of $L$ on the effectiveness of Otsu PCA-Part ($L$: \# histogram bins)}
\begin{tabular}{llrrrrr}
\hline
ID & Criterion & $L=64$ & $L=128$ & $L=256$ & $L=512$ & $L=1024$\\
\hline
\hline
9 & Initial SSE &   123527 & 123095 & \ul{122528} & 123129 & 123342\\
 & Final SSE & 118575 & 118577 & 119326 & \ul{118298} & 118616\\
\hline
10 & Initial SSE & 1855 & \ul{1807} & 1835 & 1849 & 1848\\
 & Final SSE & \ul{1742} & \ul{1742} & \ul{1742} & \ul{1742} & \ul{1742}\\
\hline
11 & Initial SSE & 2994 & 2997 & 2995 & 2995 & \ul{2991}\\
 & Final SSE & \ul{2747} & \ul{2747} & \ul{2747} & \ul{2747} & \ul{2747}\\
\hline
15 & Initial SSE & 15136 & 15117 & 15118 & \ul{15116} & 15117\\
 & Final SSE & \ul{14650} & \ul{14650} & \ul{14650} & \ul{14650} & \ul{14650}\\
\hline
\end{tabular}
\end{table}

In the remaining experiments, each random method was executed a $100$ times and statistics such as \emph{minimum}, \emph{mean}, and \emph{standard deviation} were collected for each performance criteria. The \emph{minimum} and \emph{mean} statistics represent the \emph{best} and \emph{average} case performance, respectively, while \emph{standard deviation} quantifies the \emph{variability} of performance across different runs. Note that for a deterministic method, the \emph{minimum} and \emph{mean} values are always identical and the \emph{standard deviation} is always $0$.

\begin{figure}[!ht]
\centering
 \subfigure[Minimum Initial SSE]{\includegraphics[width=0.48\columnwidth]{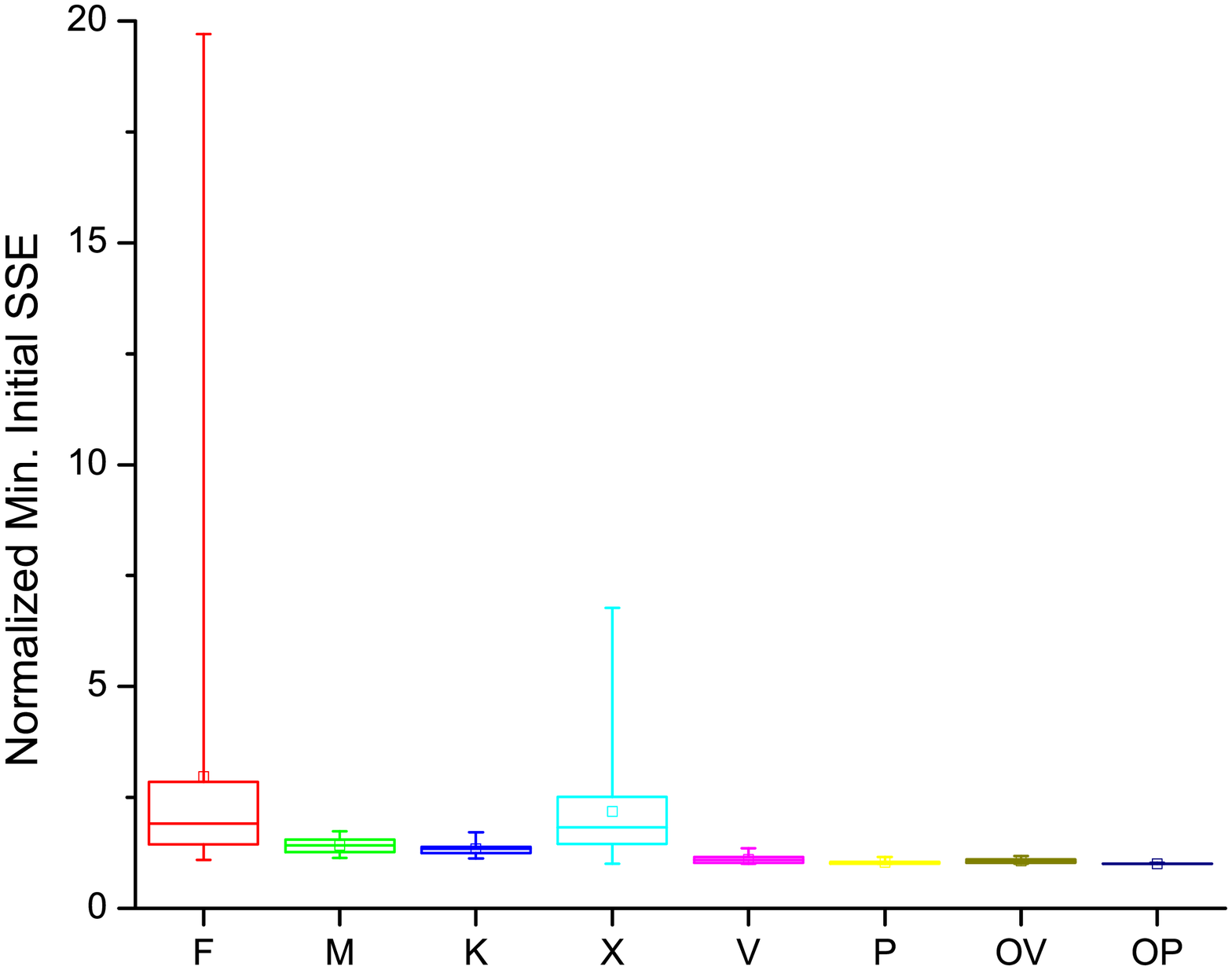}}
 \subfigure[Mean Initial SSE]{\includegraphics[width=0.48\columnwidth]{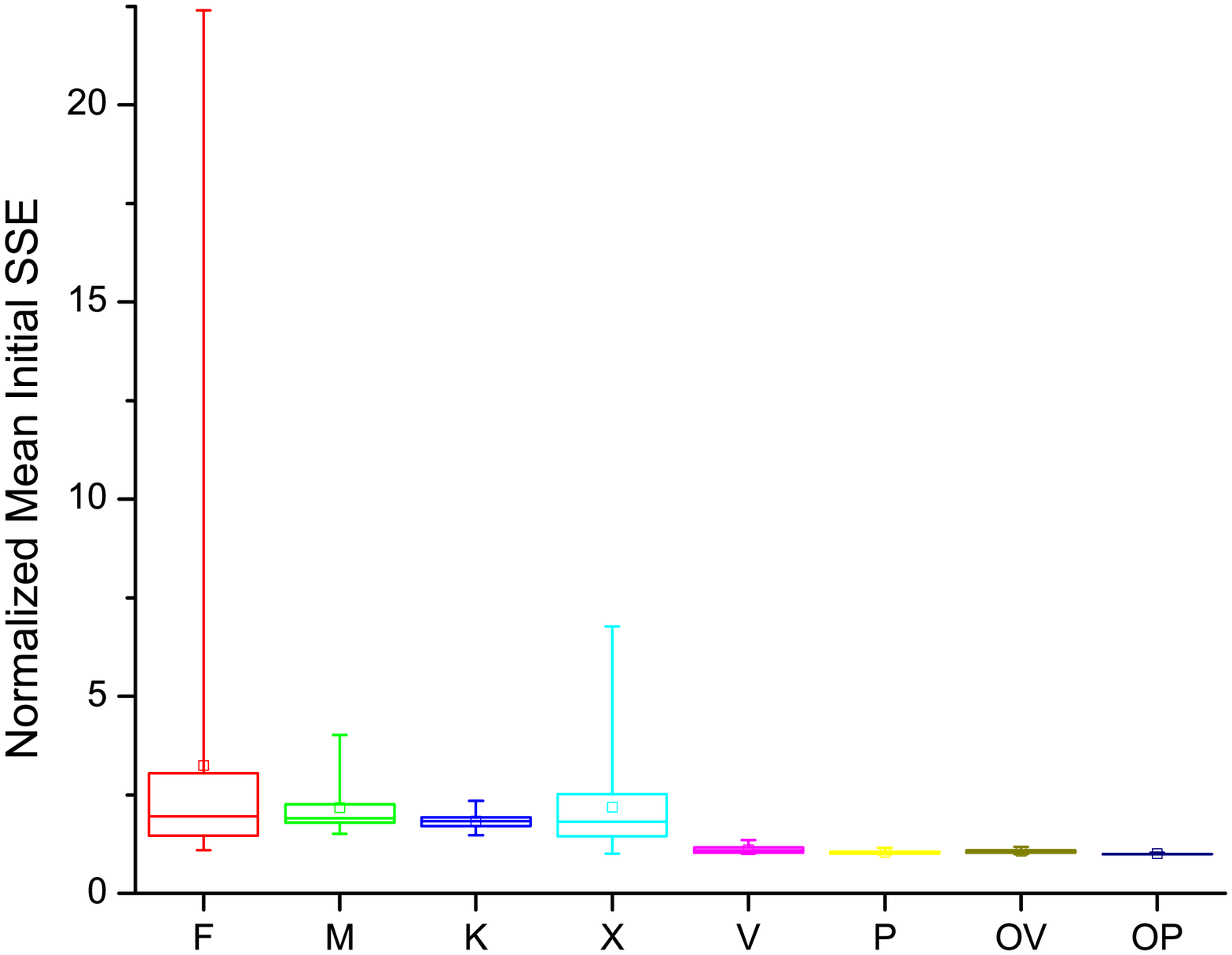}}
 \\
 \subfigure[Minimum Final SSE]{\includegraphics[width=0.48\columnwidth]{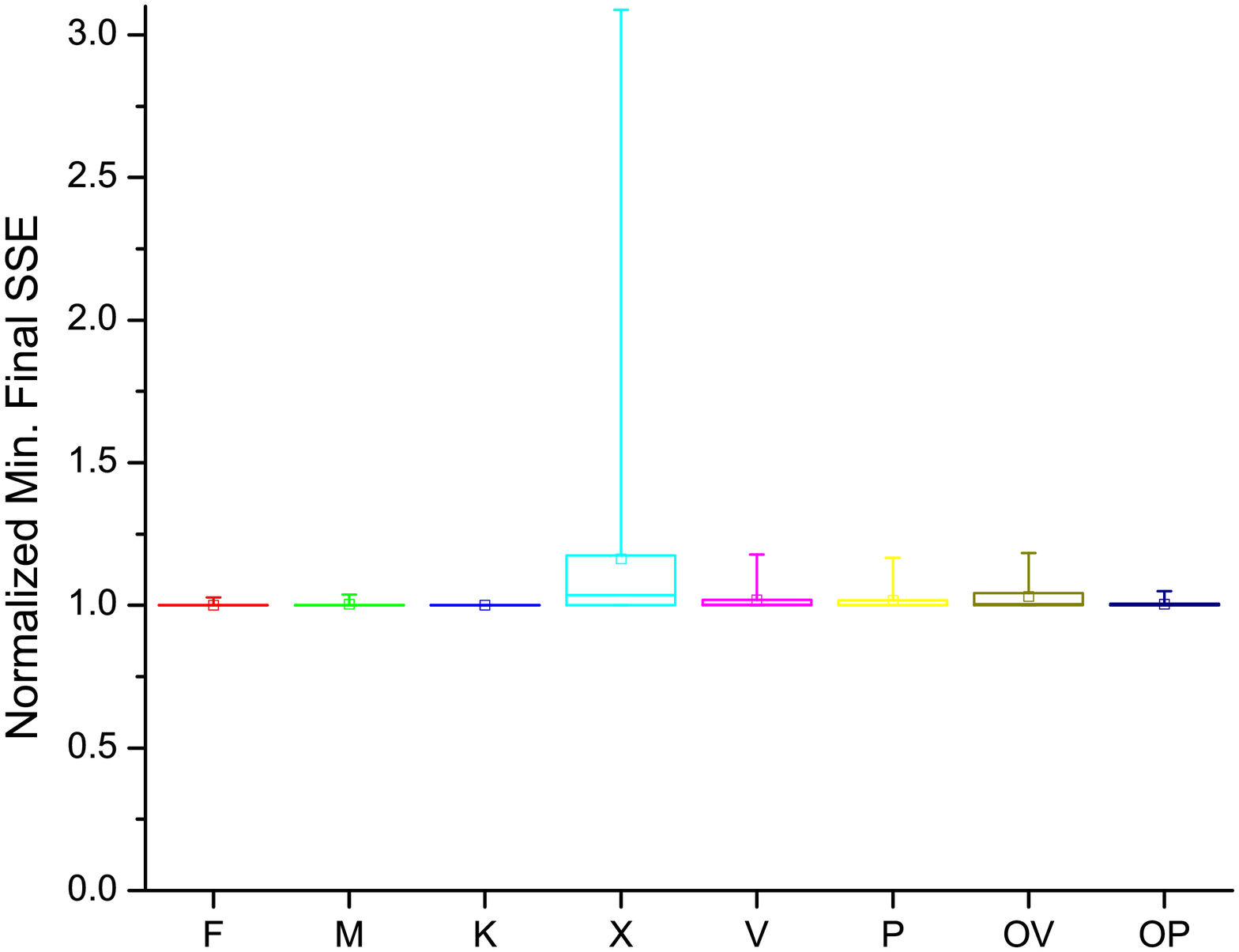}}
 \subfigure[Mean Final SSE]{\includegraphics[width=0.48\columnwidth]{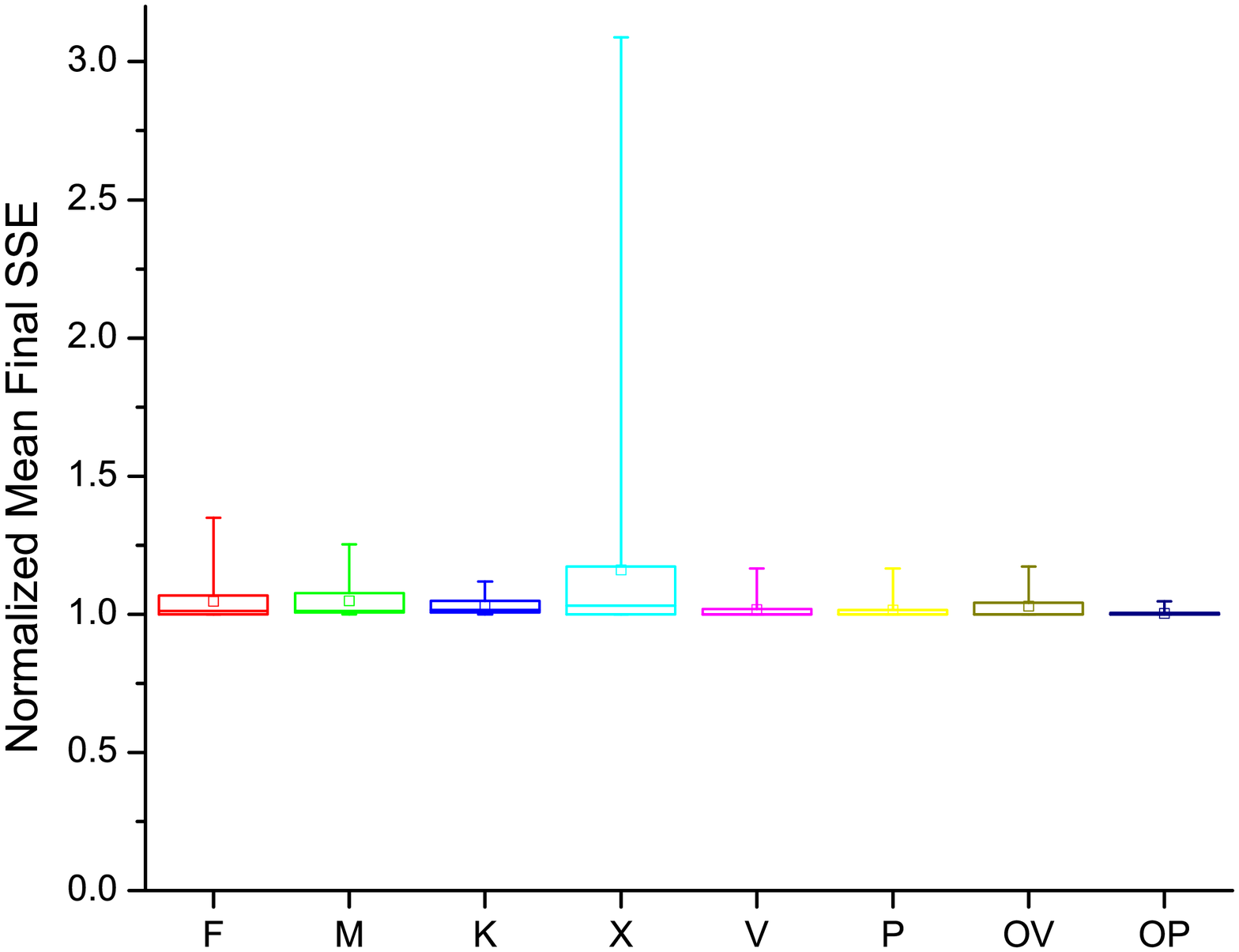}}
 \\
 \subfigure[Minimum \# Iterations]{\includegraphics[width=0.48\columnwidth]{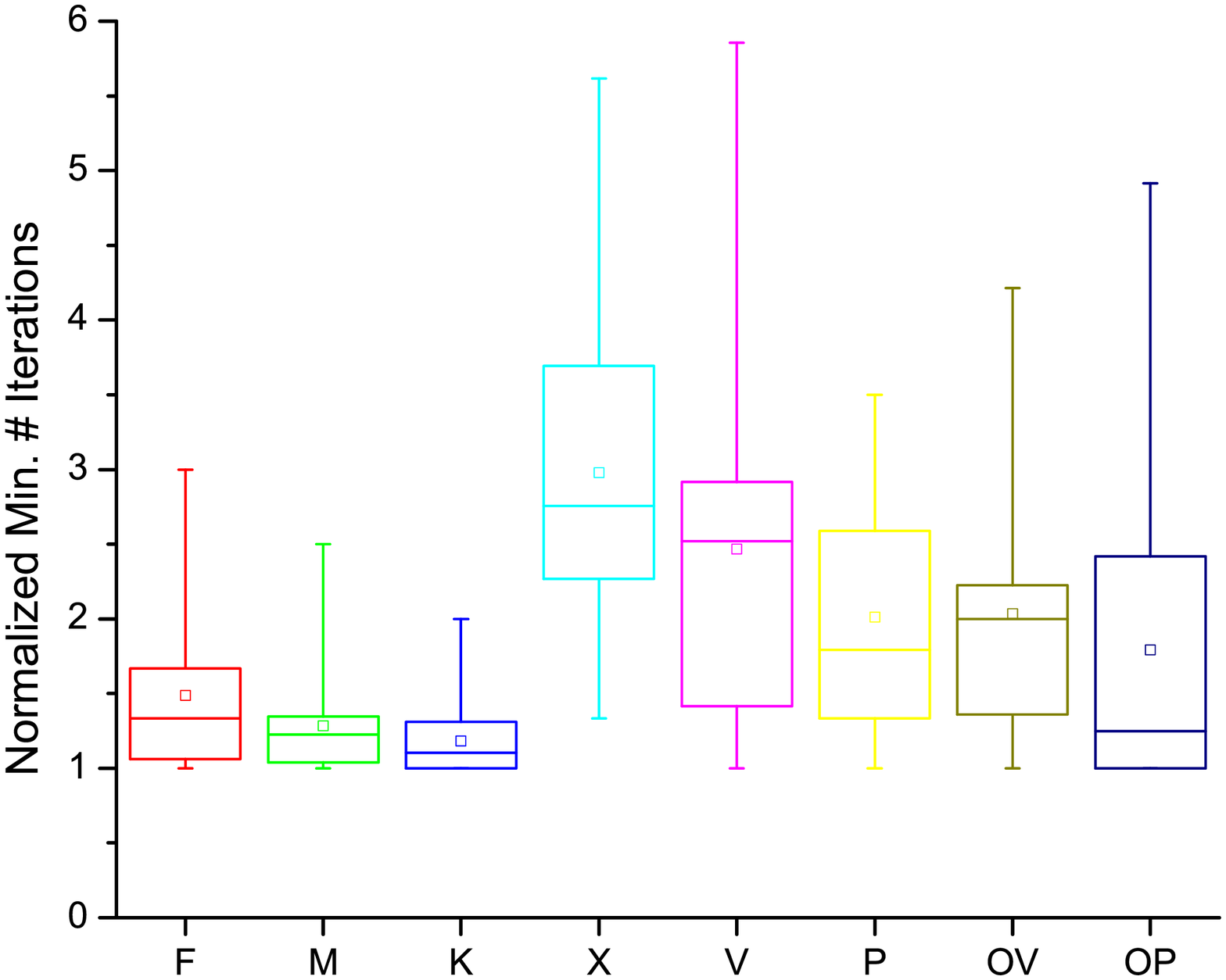}}
 \subfigure[Mean \# Iterations]{\includegraphics[width=0.48\columnwidth]{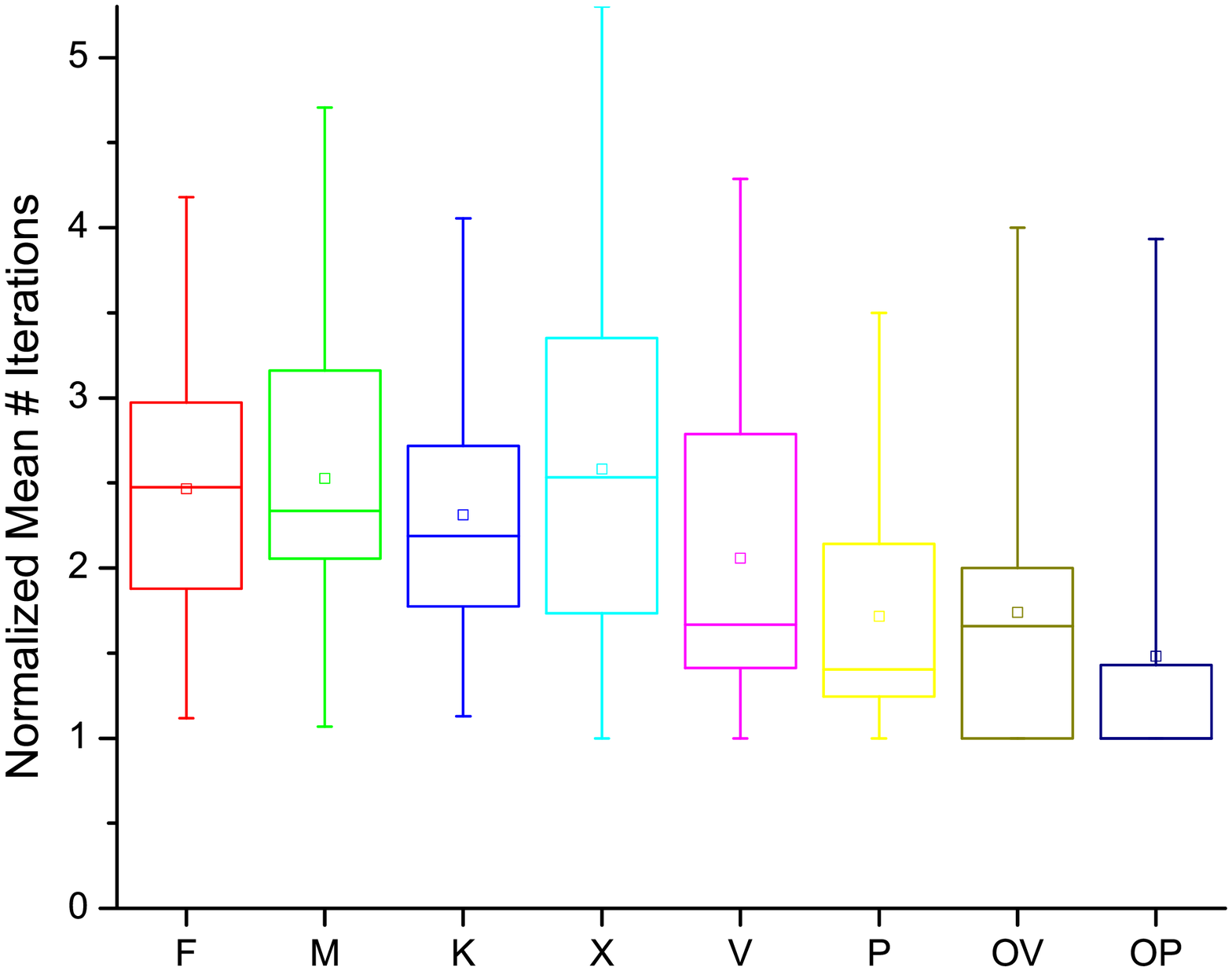}}
 \caption{Box plots of the normalized performance criteria}
 \label{fig_boxplots}
\end{figure}

Tables \ref{tab_init_sse}--\ref{tab_cpu_time} give the performance measurements for each method with respect to initial SSE, final SSE, number of iterations, and CPU time, respectively. For each of the initial SSE, final SSE, and number of iterations criteria, we calculated the ratio of the \emph{minimum}/\emph{mean}/\emph{standard deviation} value obtained by each method to the best (least) \emph{minimum}/\emph{mean}/\emph{standard deviation} value on each data set. These ratios will be henceforth referred to as the `normalized' performance criteria. For example, on the Abalone data set (1), the \emph{minimum} initial SSE of \texttt{F} is $424.92$ and the best \emph{minimum} initial SSE is $21.57$ and thus the normalized initial SSE is about $20$. This simply means that on this data set \texttt{F} obtained approximately $20$ times worse \emph{minimum} initial SSE than the best method. We then averaged\footnote{Due to outliers, the `median' statistic rather than the `mean' was used to summarize the normalized standard deviation values.} these normalized values over the $24$ data sets to quantify the \emph{overall} performance of each method with respect to each statistic (see Table \ref{tab_summary}). Note that we did not attempt to summarize the CPU time values in the same manner due to the sparsity of the data (see Table \ref{tab_cpu_time}). For convenient visualization, Figure \ref{fig_boxplots} shows box plots of the normalized performance criteria. Here, the bottom and top end of the whiskers of a box represent the \emph{minimum} and \emph{maximum}, respectively, whereas the bottom and top of the box itself are the $25$th percentile (\emph{Q1}) and $75$th percentile (\emph{Q3}), respectively. The line that passes through the box is the $50$th percentile (\emph{Q2}), i.e., the \emph{median}, while the small square inside the box denotes the \emph{mean}.

\section{Discussion}
\label{sec_disc}

\subsection{Best Case Performance Analysis}

With respect to the \emph{minimum} statistic, the following observations can be made:

\begin{itemize}
\renewcommand{\labelitemi}{$\triangleright$}
  \item {\bf Initial SSE}: \texttt{OP} is the best method, followed closely by \texttt{P}, \texttt{OV}, and \texttt{V}. On the other hand, \texttt{F} is the worst method, followed by \texttt{X}. These two methods give $2$--$3$ times worse \emph{minimum} initial SSE than the best method. It can be seen that multiple runs of random methods do not produce good initial clusterings. In contrast, only a single run of \texttt{OP}, \texttt{P}, \texttt{OV}, or \texttt{V} often gives very good results. This is because these methods are approximate clustering algorithms by themselves and thus they give reasonable results even without k-means refinement.
  \item {\bf Final SSE}: \texttt{X} is the worst method, while the remaining methods exhibit very similar performance. This homogeneity in performance is because k-means can take two disparate initial configurations to similar (or even identical) local minima. Given the abundance of local minima even in data sets of moderate size and/or dimensionality and the gradient descent nature of k-means, it is not surprising that the deterministic methods (except \texttt{X}) perform slightly worse than the random methods as the former methods were executed only once, whereas the latter ones were executed a $100$ times.
  \item {\bf Number of Iterations}: \texttt{K} is the best method, followed by \texttt{M} and \texttt{F}. \texttt{X} is the worst method. As in the case of final SSE, random methods outperform deterministic methods due to their multiple-run advantage.
\end{itemize}

\subsection{Average Case Performance Analysis}

With respect to the \emph{mean} statistic, the following observations can be made:

\begin{itemize}
\renewcommand{\labelitemi}{$\triangleright$}
  \item {\bf Initial SSE}:  \texttt{OP} is the best method, followed closely by \texttt{P}, \texttt{OV}, and \texttt{V}. The remaining methods give $1.7$--$3.2$ times worse \emph{mean} initial SSE than any of these hierarchical methods. Random methods exhibit significantly worse \emph{average} performance than the deterministic ones because the former methods can produce highly variable results across different runs (see the standard deviation values in Table \ref{tab_init_sse}).
  \item {\bf Final SSE}: This is similar to the case of \emph{minimum} final SSE, with the difference that deterministic methods (except \texttt{X}) are now slightly better than the random ones. Once again, this is because random methods can produce highly variable results due to their stochastic nature.
  \item {\bf Number of Iterations}: The ranking of the methods is similar to the case of \emph{mean} final SSE.
\end{itemize}

\subsection{Consistency Analysis}

With respect to the \emph{standard deviation} statistic, \texttt{F} is significantly better than both \texttt{M} and \texttt{K}. If, however, the application requires absolute consistency, i.e., exactly the same clustering in every run, a deterministic IM should be used.

\subsection{CPU Time Analysis}

It can be seen from Table \ref{tab_cpu_time} that, on about half of the data sets, each of the IMs require less than a few milliseconds of CPU time. On the other hand, on large and/or high-dimensional data sets efficiency differences become more prominent. It should be noted that each of the values reported in this table corresponds to a single k-means `run'. In practice, a random method is typically executed $R$ times, e.g., in this study $R=100$, and the output of the run that gives the least final SSE is taken as the result. Therefore, the total computational cost of a random method is often significantly higher than that of a deterministic method. For example, on the ISOLET data set, which has the greatest $N \times D \times K$ value among all the data sets, \texttt{K} took on the average $3,397$ milliseconds, whereas \texttt{OP} took $12,460$ milliseconds. The latter method, however, required about $27$ times less CPU time than the former one since the former was executed a total of $100$ times.

\subsection{Relative Performance Analysis}

We also determined the number of data sets (out of 24) on which \texttt{OV} and \texttt{OP} respectively performed $\{$worse than/same as/better than$\}$ \texttt{V} and \texttt{P}. Tables \ref{tab_rel_var} and \ref{tab_rel_pca} present the results for \texttt{OV} and \texttt{OP}, respectively. It can be seen that, with respect to initial SSE and number of iterations criteria, \texttt{OV} outperforms \texttt{V} more often than not. On the other hand, \texttt{OP} frequently outperforms \texttt{P} with respect to both criteria. As for final SSE, \texttt{OP} performs slightly better than \texttt{P}, whereas \texttt{OV} performs slightly worse than \texttt{V}. It appears that Otsu's method benefits \texttt{P} more than it benefits \texttt{V}. This is most likely due to the fact that histograms of projections over the major axis necessarily have a greater dynamic range and variability and thus are more amenable to thresholding compared to histograms of projections over any coordinate axis.

\subsection{Recommendations for Practitioners}

Based on the analyses presented above, the following recommendations can be made:
\begin{itemize}
\renewcommand{\labelitemi}{$\triangleright$}
	\item In general, \texttt{X} should not be used. As mentioned in Section \ref{sec_init_methods}, this method was not designed specifically as a k-means initializer \cite{Gonzalez85}. It is easy to understand and implement, but is mostly ineffective and unreliable. Furthermore, despite its low overhead, this method does not offer significant time savings since it often results in slow k-means convergence.
	\item In applications that involve small data sets, e.g., $N < 1,000$, \texttt{K} should be used. It is computationally feasible to run this method hundreds of times on such data sets given that one such run takes only a few milliseconds.
	\item In time-critical applications that involve large data sets or applications that demand determinism, the hierarchical methods should be used. These methods need to be executed only once and they lead to reasonably fast k-means convergence. The efficiency difference between \texttt{V}/\texttt{OV} and \texttt{P}/\texttt{OP} is noticeable only on high dimensional data sets such as ISOLET ($D=617$) and Musk ($D=166$). This is because \texttt{V}/\texttt{OV} calculates the direction of split by determining the coordinate axis with the greatest variance (in $\mathcal{O}(D)$ time), whereas \texttt{P}/\texttt{OP} achieves this by calculating the principal eigenvector of the cluster covariance matrix (in $\mathcal{O}(D^2)$ time using the power method \cite{Hotelling36}). Note that despite its higher computational complexity, \texttt{P}/\texttt{OP} can, in some cases, be more efficient than \texttt{V}/\texttt{OV} (see Table \ref{tab_cpu_time}). This is because the former converges significantly faster than the latter (see Table \ref{tab_summary}). The main disadvantage of these methods is that they are more complicated to implement due to their hierarchical formulation.
	\item In applications where an approximate clustering of the data set is desired, the hierarchical methods should be used. These methods produce very good initial clusterings, which makes it possible to use them as standalone clustering algorithms.
    \item Among the hierarchical methods, the ones based on PCA, i.e., \texttt{P} and \texttt{OP}, are preferable to those based on variance, i.e., \texttt{V} and \texttt{OV}. Furthermore, the proposed \texttt{OP} and \texttt{OV} methods generally outperform their respective counterparts, i.e., \texttt{P} and \texttt{V}, especially with respect to initial SSE and number of iterations.
\end{itemize}

\section{Conclusions}
\label{sec_conc}
In this paper, we presented a simple modification to Var-Part and PCA-Part, two hierarchical k-means initialization methods that are linear, deterministic, and order-invariant. We compared the original methods and their modified versions to some of the most popular linear initialization methods, namely Forgy's method, Macqueen's second method, maximin, and k-means++, on a large and diverse collection of data sets from the UCI Machine Learning Repository. The results demonstrated that, despite their deterministic nature, Var-Part and PCA-Part are highly competitive with one of the best random initialization methods to date, i.e., k-means++. In addition, the proposed modification significantly improves the performance of both hierarchical methods. The presented Var-Part and PCA-Part variants can be used to initialize k-means effectively, particularly in time-critical applications that involve large data sets. Alternatively, they can be used as approximate clustering algorithms without additional k-means refinement.

\section{Acknowledgments}

This publication was made possible by grants from the Louisiana Board of Regents (LEQSF2008-11-RD-A-12) and National Science Foundation (0959583, 1117457).

\newpage

\begin{center}
{
\tiny
\tablecaption{\label{tab_init_sse} Initial SSE comparison of the initialization methods}
\begin{xtabular}{c|c|r|r|r|r|r|r|r|r}
\hline
 & & \texttt{F} & \texttt{M} & \texttt{K} & \texttt{X} & \texttt{V} & \texttt{P} & \texttt{OV} & \texttt{OP}\\
\hline
\hline
\multirow{2}{*}{1} & min & 425 & 33 & 29 & 95 & 24 & 23 & 23 & \ul{22}\\
& mean & 483 $\pm$ 20 & 46 $\pm$ 10 & 34 $\pm$ 2 & 95 $\pm$ 0 & 24 $\pm$ 0 & 23 $\pm$ 0 & 23 $\pm$ 0 & \ul{22} $\pm$ 0\\
\hline
\multirow{2}{*}{2} & min & 534 & 318 & 304 & 498 & 247 & 240 & 258 & \ul{239}\\
& mean & 575 $\pm$ 15 & 706 $\pm$ 354 & 560 $\pm$ 349 & 498 $\pm$ 0 & 247 $\pm$ 0 & 240 $\pm$ 0 & 258 $\pm$ 0 & \ul{239} $\pm$ 0\\
\hline
\multirow{2}{*}{3} & min & 20 & 11 & 9 & 19 & 8 & 8 & 8 & \ul{7}\\
& mean & 27 $\pm$ 3 & 20 $\pm$ 8 & 13 $\pm$ 2 & 19 $\pm$ 0 & 8 $\pm$ 0 & 8 $\pm$ 0 & 8 $\pm$ 0 & \ul{7} $\pm$ 0\\
\hline
\multirow{2}{*}{4} & min & 54 & 26 & 26 & 48 & 20 & \ul{19} & \ul{19} & 20\\
& mean & 61 $\pm$ 2 & 40 $\pm$ 7 & 33 $\pm$ 5 & 48 $\pm$ 0 & 20 $\pm$ 0 & \ul{19} $\pm$ 0 & \ul{19} $\pm$ 0 & 20 $\pm$ 0\\
\hline
\multirow{2}{*}{5} & min & 42 & 24 & 25 & 45 & 21 & 20 & 21 & \ul{18}\\
& mean & 48 $\pm$ 2 & 40 $\pm$ 9 & 32 $\pm$ 5 & 45 $\pm$ 0 & 21 $\pm$ 0 & 20 $\pm$ 0 & 21 $\pm$ 0 & \ul{18} $\pm$ 0\\
\hline
\multirow{2}{*}{6} & min & 372 & 361 & 341 & 409 & 249 & 250 & 249 & \ul{244}\\
& mean & 396 $\pm$ 8 & 463 $\pm$ 58 & 450 $\pm$ 49 & 409 $\pm$ 0 & 249 $\pm$ 0 & 250 $\pm$ 0 & 249 $\pm$ 0 & \ul{244} $\pm$ 0\\
\hline
\multirow{2}{*}{7} & min & 771 & 749 & 720 & 827 & 632 & \ul{629} & 636 & \ul{629}\\
& mean & 814 $\pm$ 12 & 1246 $\pm$ 463 & 1237 $\pm$ 468 & 827 $\pm$ 0 & 632 $\pm$ 0 & \ul{629} $\pm$ 0 & 636 $\pm$ 0 & \ul{629} $\pm$ 0\\
\hline
\multirow{2}{*}{8} & min & 26 & 9 & 9 & 18 & 8 & 8 & \ul{7} & \ul{7}\\
& mean & 34 $\pm$ 4 & 28 $\pm$ 23 & 16 $\pm$ 6 & 18 $\pm$ 0 & 8 $\pm$ 0 & 8 $\pm$ 0 & \ul{7} $\pm$ 0 & \ul{7} $\pm$ 0\\
\hline
\multirow{2}{*}{9} & min & 218965 & 212238 & 210387 & 221163 & 145444 & 124958 & 144658 & \ul{122528}\\
& mean & 223003 $\pm$ 1406 & 224579 $\pm$ 5416 & 223177 $\pm$ 4953 & 221163 $\pm$ 0 & 145444 $\pm$ 0 & 124958 $\pm$ 0 & 144658 $\pm$ 0 & \ul{122528} $\pm$ 0\\
\hline
\multirow{2}{*}{10} & min & 7763 & 2637 & 2458 & 4816 & 2050 & 2116 & 1919 & \ul{1835}\\
& mean & 8057 $\pm$ 98 & 4825 $\pm$ 1432 & 3561 $\pm$ 747 & 4816 $\pm$ 0 & 2050 $\pm$ 0 & 2116 $\pm$ 0 & 1919 $\pm$ 0 & \ul{1835} $\pm$ 0\\
\hline
\multirow{2}{*}{11} & min & 7100 & 4203 & 4158 & 5632 & 3456 & 3101 & 3231 & \ul{2995}\\
& mean & 7225 $\pm$ 30 & 4532 $\pm$ 165 & 4501 $\pm$ 176 & 5632 $\pm$ 0 & 3456 $\pm$ 0 & 3101 $\pm$ 0 & 3231 $\pm$ 0 & \ul{2995} $\pm$ 0\\
\hline
\multirow{2}{*}{12} & min & 4343 & 3348 & 3296 & 4361 & 3056 & 2927 & 3060 & \ul{2923}\\
& mean & 4392 $\pm$ 13 & 5525 $\pm$ 1816 & 5346 $\pm$ 1672 & 4361 $\pm$ 0 & 3056 $\pm$ 0 & 2927 $\pm$ 0 & 3060 $\pm$ 0 & \ul{2923} $\pm$ 0\\
\hline
\multirow{2}{*}{13} & min & 4416 & 5205 & 5247 & 4485 & 3354 & 3266 & 3315 & \ul{3180}\\
& mean & 4475 $\pm$ 25 & 5693 $\pm$ 315 & 5758 $\pm$ 283 & 4485 $\pm$ 0 & 3354 $\pm$ 0 & 3266 $\pm$ 0 & 3315 $\pm$ 0 & \ul{3180} $\pm$ 0\\
\hline
\multirow{2}{*}{14} & min & 53508 & 56841 & 56822 & 54629 & 37334 & 37142 & 37282 & \ul{36375}\\
& mean & 54312 $\pm$ 244 & 82411 $\pm$ 14943 & 75532 $\pm$ 12276 & 54629 $\pm$ 0 & 37334 $\pm$ 0 & 37142 $\pm$ 0 & 37282 $\pm$ 0 & \ul{36375} $\pm$ 0\\
\hline
\multirow{2}{*}{15} & min & 25466 & 25492 & 24404 & 25291 & 17476 & 15714 & 17504 & \ul{15118}\\
& mean & 25811 $\pm$ 99 & 28596 $\pm$ 1550 & 27614 $\pm$ 1499 & 25291 $\pm$ 0 & 17476 $\pm$ 0 & 15714 $\pm$ 0 & 17504 $\pm$ 0 & \ul{15118} $\pm$ 0\\
\hline
\multirow{2}{*}{16} & min & 633 & 275 & 250 & 635 & 300 & 230 & 232 & \ul{222}\\
& mean & 648 $\pm$ 6 & 423 $\pm$ 74 & 372 $\pm$ 72 & 635 $\pm$ 0 & 300 $\pm$ 0 & 230 $\pm$ 0 & 232 $\pm$ 0 & \ul{222} $\pm$ 0\\
\hline
\multirow{2}{*}{17} & min & 152 & 144 & 141 & 156 & 124 & 122 & 123 & \ul{121}\\
& mean & 156 $\pm$ 1 & 216 $\pm$ 44 & 219 $\pm$ 61 & 156 $\pm$ 0 & 124 $\pm$ 0 & 122 $\pm$ 0 & 123 $\pm$ 0 & \ul{121} $\pm$ 0\\
\hline
\multirow{2}{*}{18} & min & 1788 & 438 & 328 & 1818 & 316 & 309 & 276 & \ul{268}\\
& mean & 1806 $\pm$ 6 & 946 $\pm$ 290 & 494 $\pm$ 115 & 1818 $\pm$ 0 & 316 $\pm$ 0 & 309 $\pm$ 0 & 276 $\pm$ 0 & \ul{268} $\pm$ 0\\
\hline
\multirow{2}{*}{19} & min & 834 & 873 & 881 & 772 & 782 & 783 & 792 & \ul{765}\\
& mean & 838 $\pm$ 1 & 1186 $\pm$ 386 & 1124 $\pm$ 244 & 772 $\pm$ 0 & 782 $\pm$ 0 & 783 $\pm$ 0 & 792 $\pm$ 0 & \ul{765} $\pm$ 0\\
\hline
\multirow{2}{*}{20} & min & 269 & 295 & 297 & 277 & 232 & 222 & 225 & \ul{214}\\
& mean & 281 $\pm$ 4 & 384 $\pm$ 88 & 413 $\pm$ 159 & 277 $\pm$ 0 & 232 $\pm$ 0 & 222 $\pm$ 0 & 225 $\pm$ 0 & \ul{214} $\pm$ 0\\
\hline
\multirow{2}{*}{21} & min & 10976 & 11834 & 11829 & 11004 & 8517 & 7805 & 8706 & \ul{7802}\\
& mean & 11082 $\pm$ 34 & 14814 $\pm$ 1496 & 14435 $\pm$ 1276 & 11004 $\pm$ 0 & 8517 $\pm$ 0 & 7805 $\pm$ 0 & 8706 $\pm$ 0 & \ul{7802} $\pm$ 0\\
\hline
\multirow{2}{*}{22} & min & 719 & 473 & 449 & 733 & 386 & 361 & 364 & \ul{351}\\
& mean & 729 $\pm$ 4 & 601 $\pm$ 59 & 567 $\pm$ 64 & 733 $\pm$ 0 & 386 $\pm$ 0 & 361 $\pm$ 0 & 364 $\pm$ 0 & \ul{351} $\pm$ 0\\
\hline
\multirow{2}{*}{23} & min & 78 & 76 & 70 & 87 & 51 & 53 & \ul{50} & 51\\
& mean & 87 $\pm$ 3 & 113 $\pm$ 22 & 101 $\pm$ 20 & 87 $\pm$ 0 & 51 $\pm$ 0 & 53 $\pm$ 0 & \ul{50} $\pm$ 0 & 51 $\pm$ 0\\
\hline
\multirow{2}{*}{24} & min & 144 & 89 & 83 & 115 & 77 & \ul{63} & 73 & \ul{63}\\
& mean & 149 $\pm$ 2 & 110 $\pm$ 8 & 101 $\pm$ 9 & 115 $\pm$ 0 & 77 $\pm$ 0 & \ul{63} $\pm$ 0 & 73 $\pm$ 0 & \ul{63} $\pm$ 0\\
\hline
\end{xtabular}
}
\end{center}

\newpage

\begin{center}
{
\tiny
\tablecaption{\label{tab_final_sse} Final SSE comparison of the initialization methods}
\begin{xtabular}{c|c|r|r|r|r|r|r|r|r}
\hline
 & & \texttt{F} & \texttt{M} & \texttt{K} & \texttt{X} & \texttt{V} & \texttt{P} & \texttt{OV} & \texttt{OP}\\
\hline
\hline
\multirow{2}{*}{1} & min & \ul{21} & 22 & \ul{21} & 25 & \ul{21} & \ul{21} & \ul{21} & \ul{21}\\
& mean & 23 $\pm$ 1 & 22 $\pm$ 1 & 22 $\pm$ 0 & 25 $\pm$ 0 & \ul{21} $\pm$ 0 & \ul{21} $\pm$ 0 & \ul{21} $\pm$ 0 & \ul{21} $\pm$ 0\\
\hline
\multirow{2}{*}{2} & min & \ul{239} & \ul{239} & \ul{239} & \ul{239} & \ul{239} & \ul{239} & \ul{239} & \ul{239}\\
& mean & \ul{239} $\pm$ 0 & \ul{239} $\pm$ 0 & \ul{239} $\pm$ 0 & \ul{239} $\pm$ 0 & \ul{239} $\pm$ 0 & \ul{239} $\pm$ 0 & \ul{239} $\pm$ 0 & \ul{239} $\pm$ 0\\
\hline
\multirow{2}{*}{3} & min & \ul{7} & \ul{7} & \ul{7} & \ul{7} & \ul{7} & \ul{7} & 8 & \ul{7}\\
& mean & 8 $\pm$ 1 & 9 $\pm$ 1 & 8 $\pm$ 1 & \ul{7} $\pm$ 0 & \ul{7} $\pm$ 0 & \ul{7} $\pm$ 0 & 8 $\pm$ 0 & \ul{7} $\pm$ 0\\
\hline
\multirow{2}{*}{4} & min & \ul{17} & \ul{17} & \ul{17} & 19 & \ul{17} & 18 & 18 & 18\\
& mean & 19 $\pm$ 1 & 19 $\pm$ 2 & 19 $\pm$ 1 & 19 $\pm$ 0 & \ul{17} $\pm$ 0 & 18 $\pm$ 0 & 18 $\pm$ 0 & 18 $\pm$ 0\\
\hline
\multirow{2}{*}{5} & min & \ul{18} & \ul{18} & \ul{18} & 23 & 19 & 19 & 20 & \ul{18}\\
& mean & 20 $\pm$ 1 & 21 $\pm$ 2 & 20 $\pm$ 2 & 23 $\pm$ 0 & 19 $\pm$ 0 & 19 $\pm$ 0 & 20 $\pm$ 0 & \ul{18} $\pm$ 0\\
\hline
\multirow{2}{*}{6} & min & \ul{243} & \ul{243} & \ul{243} & 249 & 248 & \ul{243} & 248 & \ul{243}\\
& mean & 252 $\pm$ 8 & 252 $\pm$ 8 & 252 $\pm$ 8 & 249 $\pm$ 0 & 248 $\pm$ 0 & \ul{243} $\pm$ 0 & 248 $\pm$ 0 & \ul{243} $\pm$ 0\\
\hline
\multirow{2}{*}{7} & min & \ul{629} & \ul{629} & \ul{629} & 826 & \ul{629} & \ul{629} & \ul{629} & \ul{629}\\
& mean & \ul{629} $\pm$ 0 & 643 $\pm$ 50 & 641 $\pm$ 47 & 826 $\pm$ 0 & \ul{629} $\pm$ 0 & \ul{629} $\pm$ 0 & \ul{629} $\pm$ 0 & \ul{629} $\pm$ 0\\
\hline
\multirow{2}{*}{8} & min & \ul{7} & \ul{7} & \ul{7} & \ul{7} & \ul{7} & \ul{7} & \ul{7} & \ul{7}\\
& mean & 8 $\pm$ 1 & 8 $\pm$ 2 & \ul{7} $\pm$ 1 & \ul{7} $\pm$ 0 & \ul{7} $\pm$ 0 & \ul{7} $\pm$ 0 & \ul{7} $\pm$ 0 & \ul{7} $\pm$ 0\\
\hline
\multirow{2}{*}{9} & min & 117872 & 117764 & \ul{117710} & 135818 & 118495 & 118386 & 118033 & 119326\\
& mean & 119650 $\pm$ 945 & 119625 $\pm$ 947 & 119536 $\pm$ 934 & 135818 $\pm$ 0 & 118495 $\pm$ 0 & 118386 $\pm$ 0 & \ul{118033} $\pm$ 0 & 119326 $\pm$ 0\\
\hline
\multirow{2}{*}{10} & min & \ul{1742} & \ul{1742} & \ul{1742} & \ul{1742} & \ul{1742} & \ul{1742} & \ul{1742} & \ul{1742}\\
& mean & \ul{1742} $\pm$ 0 & \ul{1742} $\pm$ 0 & 1744 $\pm$ 28 & \ul{1742} $\pm$ 0 & \ul{1742} $\pm$ 0 & \ul{1742} $\pm$ 0 & \ul{1742} $\pm$ 0 & \ul{1742} $\pm$ 0\\
\hline
\multirow{2}{*}{11} & min & 2723 & 2718 & \ul{2716} & 2749 & 2735 & 2745 & 2734 & 2747\\
& mean & 2772 $\pm$ 29 & 2757 $\pm$ 19 & 2751 $\pm$ 19 & 2749 $\pm$ 0 & 2735 $\pm$ 0 & 2745 $\pm$ 0 & \ul{2734} $\pm$ 0 & 2747 $\pm$ 0\\
\hline
\multirow{2}{*}{12} & min & \ul{2923} & \ul{2923} & \ul{2923} & \ul{2923} & \ul{2923} & \ul{2923} & \ul{2923} & \ul{2923}\\
& mean & \ul{2923} $\pm$ 0 & \ul{2923} $\pm$ 0 & \ul{2923} $\pm$ 0 & \ul{2923} $\pm$ 0 & \ul{2923} $\pm$ 0 & \ul{2923} $\pm$ 0 & \ul{2923} $\pm$ 0 & \ul{2923} $\pm$ 0\\
\hline
\multirow{2}{*}{13} & min & \ul{3127} & 3128 & 3128 & 3316 & 3137 & 3214 & 3143 & 3153\\
& mean & 3166 $\pm$ 31 & 3172 $\pm$ 29 & 3173 $\pm$ 35 & 3316 $\pm$ 0 & \ul{3137} $\pm$ 0 & 3214 $\pm$ 0 & 3143 $\pm$ 0 & 3153 $\pm$ 0\\
\hline
\multirow{2}{*}{14} & min & \ul{36373} & \ul{36373} & \ul{36373} & \ul{36373} & \ul{36373} & \ul{36373} & \ul{36373} & \ul{36373}\\
& mean & 37296 $\pm$ 1902 & 37163 $\pm$ 1338 & 37058 $\pm$ 1626 & \ul{36373} $\pm$ 0 & \ul{36373} $\pm$ 0 & \ul{36373} $\pm$ 0 & \ul{36373} $\pm$ 0 & \ul{36373} $\pm$ 0\\
\hline
\multirow{2}{*}{15} & min & \ul{14559} & \ul{14559} & \ul{14559} & 14679 & 14581 & 14807 & 14581 & 14650\\
& mean & 14687 $\pm$ 216 & 14752 $\pm$ 236 & 14747 $\pm$ 245 & 14679 $\pm$ 0 & \ul{14581} $\pm$ 0 & 14807 $\pm$ 0 & \ul{14581} $\pm$ 0 & 14650 $\pm$ 0\\
\hline
\multirow{2}{*}{16} & min & \ul{215} & \ul{215} & \ul{215} & 230 & 227 & \ul{215} & 229 & 216\\
& mean & 217 $\pm$ 4 & 216 $\pm$ 2 & 220 $\pm$ 7 & 230 $\pm$ 0 & 227 $\pm$ 0 & \ul{215} $\pm$ 0 & 229 $\pm$ 0 & 216 $\pm$ 0\\
\hline
\multirow{2}{*}{17} & min & \ul{121} & \ul{121} & \ul{121} & \ul{121} & \ul{121} & \ul{121} & \ul{121} & \ul{121}\\
& mean & \ul{121} $\pm$ 0 & 122 $\pm$ 5 & 122 $\pm$ 5 & \ul{121} $\pm$ 0 & \ul{121} $\pm$ 0 & \ul{121} $\pm$ 0 & \ul{121} $\pm$ 0 & \ul{121} $\pm$ 0\\
\hline
\multirow{2}{*}{18} & min & \ul{235} & \ul{235} & \ul{235} & 726 & \ul{235} & 274 & 274 & \ul{235}\\
& mean & 317 $\pm$ 46 & 272 $\pm$ 23 & 260 $\pm$ 31 & 726 $\pm$ 0 & \ul{235} $\pm$ 0 & 274 $\pm$ 0 & 274 $\pm$ 0 & \ul{235} $\pm$ 0\\
\hline
\multirow{2}{*}{19} & min & \ul{765} & \ul{765} & \ul{765} & \ul{765} & 778 & 778 & 778 & \ul{765}\\
& mean & 778 $\pm$ 3 & 779 $\pm$ 14 & 785 $\pm$ 19 & \ul{765} $\pm$ 0 & 778 $\pm$ 0 & 778 $\pm$ 0 & 778 $\pm$ 0 & \ul{765} $\pm$ 0\\
\hline
\multirow{2}{*}{20} & min & \ul{214} & \ul{214} & \ul{214} & \ul{214} & \ul{214} & \ul{214} & \ul{214} & \ul{214}\\
& mean & \ul{214} $\pm$ 0 & 215 $\pm$ 5 & \ul{214} $\pm$ 0 & \ul{214} $\pm$ 0 & \ul{214} $\pm$ 0 & \ul{214} $\pm$ 0 & \ul{214} $\pm$ 0 & \ul{214} $\pm$ 0\\
\hline
\multirow{2}{*}{21} & min & \ul{7772} & \ul{7772} & \ul{7772} & \ul{7772} & 7774 & 7774 & 7774 & \ul{7772}\\
& mean & 7799 $\pm$ 93 & 7821 $\pm$ 124 & 7831 $\pm$ 140 & \ul{7772} $\pm$ 0 & 7774 $\pm$ 0 & 7774 $\pm$ 0 & 7774 $\pm$ 0 & \ul{7772} $\pm$ 0\\
\hline
\multirow{2}{*}{22} & min & \ul{334} & \ul{334} & \ul{334} & 399 & 335 & \ul{334} & 335 & 335\\
& mean & 335 $\pm$ 1 & 336 $\pm$ 3 & 336 $\pm$ 3 & 399 $\pm$ 0 & 335 $\pm$ 0 & \ul{334} $\pm$ 0 & 335 $\pm$ 0 & 335 $\pm$ 0\\
\hline
\multirow{2}{*}{23} & min & \ul{49} & \ul{49} & \ul{49} & 63 & \ul{49} & \ul{49} & \ul{49} & \ul{49}\\
& mean & \ul{49} $\pm$ 0 & \ul{49} $\pm$ 2 & \ul{49} $\pm$ 2 & 63 $\pm$ 0 & \ul{49} $\pm$ 0 & \ul{49} $\pm$ 0 & \ul{49} $\pm$ 0 & \ul{49} $\pm$ 0\\
\hline
\multirow{2}{*}{24} & min & \ul{58} & \ul{58} & \ul{58} & 61 & 69 & 59 & 69 & 59\\
& mean & 64 $\pm$ 6 & 69 $\pm$ 6 & 63 $\pm$ 5 & 61 $\pm$ 0 & 69 $\pm$ 0 & \ul{59} $\pm$ 0 & 69 $\pm$ 0 & \ul{59} $\pm$ 0\\
\hline
\end{xtabular}
}
\end{center}

\newpage

\begin{center}
{
\tiny
\tablecaption{\label{tab_num_iters} Number of iterations comparison of the initialization methods}
\begin{xtabular}{c|c|r|r|r|r|r|r|r|r}
\hline
 & & \texttt{F} & \texttt{M} & \texttt{K} & \texttt{X} & \texttt{V} & \texttt{P} & \texttt{OV} & \texttt{OP}\\
\hline
\hline
\multirow{2}{*}{1} & min & 59 & 29 & \ul{22} & 100 & 50 & 43 & 31 & 38\\
& mean & 90 $\pm$ 11 & 68 $\pm$ 19 & 48 $\pm$ 17 & 100 $\pm$ 0 & 50 $\pm$ 0 & 43 $\pm$ 0 & \ul{31} $\pm$ 0 & 38 $\pm$ 0\\
\hline
\multirow{2}{*}{2} & min & 4 & 4 & 4 & 8 & 4 & 4 & 5 & \ul{3}\\
& mean & 5 $\pm$ 0 & 6 $\pm$ 1 & 6 $\pm$ 1 & 8 $\pm$ 0 & 4 $\pm$ 0 & 4 $\pm$ 0 & 5 $\pm$ 0 & \ul{3} $\pm$ 0\\
\hline
\multirow{2}{*}{3} & min & 5 & 5 & \ul{3} & 7 & 6 & 7 & 5 & \ul{3}\\
& mean & 10 $\pm$ 2 & 9 $\pm$ 3 & 7 $\pm$ 2 & 7 $\pm$ 0 & 6 $\pm$ 0 & 7 $\pm$ 0 & 5 $\pm$ 0 & \ul{3} $\pm$ 0\\
\hline
\multirow{2}{*}{4} & min & 8 & \ul{6} & 7 & 14 & 17 & 7 & 12 & \ul{6}\\
& mean & 15 $\pm$ 6 & 15 $\pm$ 5 & 14 $\pm$ 5 & 14 $\pm$ 0 & 17 $\pm$ 0 & 7 $\pm$ 0 & 12 $\pm$ 0 & \ul{6} $\pm$ 0\\
\hline
\multirow{2}{*}{5} & min & 6 & 5 & \ul{4} & 6 & 6 & 5 & 9 & \ul{4}\\
& mean & 10 $\pm$ 3 & 11 $\pm$ 4 & 9 $\pm$ 3 & 6 $\pm$ 0 & 6 $\pm$ 0 & 5 $\pm$ 0 & 9 $\pm$ 0 & \ul{4} $\pm$ 0\\
\hline
\multirow{2}{*}{6} & min & 5 & 5 & 5 & 12 & \ul{3} & 4 & \ul{3} & 4\\
& mean & 11 $\pm$ 3 & 10 $\pm$ 3 & 9 $\pm$ 3 & 12 $\pm$ 0 & \ul{3} $\pm$ 0 & 4 $\pm$ 0 & \ul{3} $\pm$ 0 & 4 $\pm$ 0\\
\hline
\multirow{2}{*}{7} & min & 4 & 3 & 3 & 3 & 3 & 3 & 4 & \ul{2}\\
& mean & 5 $\pm$ 1 & 7 $\pm$ 2 & 8 $\pm$ 2 & 3 $\pm$ 0 & 3 $\pm$ 0 & 3 $\pm$ 0 & 4 $\pm$ 0 & \ul{2} $\pm$ 0\\
\hline
\multirow{2}{*}{8} & min & 4 & 4 & \ul{3} & 6 & 4 & 4 & 6 & \ul{3}\\
& mean & 9 $\pm$ 3 & 8 $\pm$ 2 & 7 $\pm$ 3 & 6 $\pm$ 0 & 4 $\pm$ 0 & 4 $\pm$ 0 & 6 $\pm$ 0 & \ul{3} $\pm$ 0\\
\hline
\multirow{2}{*}{9} & min & 18 & 19 & \ul{14} & 32 & 82 & 45 & 59 & 39\\
& mean & 43 $\pm$ 15 & 40 $\pm$ 14 & 36 $\pm$ 13 & \ul{32} $\pm$ 0 & 82 $\pm$ 0 & 45 $\pm$ 0 & 59 $\pm$ 0 & 39 $\pm$ 0\\
\hline
\multirow{2}{*}{10} & min & 12 & 12 & 11 & 53 & 28 & 27 & \ul{10} & 23\\
& mean & 28 $\pm$ 8 & 33 $\pm$ 10 & 29 $\pm$ 9 & 53 $\pm$ 0 & 28 $\pm$ 0 & 27 $\pm$ 0 & \ul{10} $\pm$ 0 & 23 $\pm$ 0\\
\hline
\multirow{2}{*}{11} & min & 39 & 37 & \ul{31} & 73 & 100 & 83 & 67 & 85\\
& mean & 75 $\pm$ 19 & 72 $\pm$ 18 & 76 $\pm$ 18 & 73 $\pm$ 0 & 100 $\pm$ 0 & 83 $\pm$ 0 & \ul{67} $\pm$ 0 & 85 $\pm$ 0\\
\hline
\multirow{2}{*}{12} & min & \ul{9} & 10 & 10 & 35 & 25 & 10 & 26 & \ul{9}\\
& mean & 18 $\pm$ 5 & 18 $\pm$ 5 & 20 $\pm$ 6 & 35 $\pm$ 0 & 25 $\pm$ 0 & 10 $\pm$ 0 & 26 $\pm$ 0 & \ul{9} $\pm$ 0\\
\hline
\multirow{2}{*}{13} & min & \ul{13} & 14 & \ul{13} & 37 & 14 & 25 & 17 & \ul{13}\\
& mean & 29 $\pm$ 10 & 30 $\pm$ 10 & 30 $\pm$ 11 & 37 $\pm$ 0 & 14 $\pm$ 0 & 25 $\pm$ 0 & 17 $\pm$ 0 & \ul{13} $\pm$ 0\\
\hline
\multirow{2}{*}{14} & min & 4 & 4 & 4 & 8 & 5 & 5 & 5 & \ul{3}\\
& mean & 6 $\pm$ 1 & 6 $\pm$ 1 & 6 $\pm$ 1 & 8 $\pm$ 0 & 5 $\pm$ 0 & 5 $\pm$ 0 & 5 $\pm$ 0 & \ul{3} $\pm$ 0\\
\hline
\multirow{2}{*}{15} & min & \ul{12} & \ul{12} & 14 & 36 & 16 & 22 & 15 & 59\\
& mean & 31 $\pm$ 13 & 33 $\pm$ 14 & 30 $\pm$ 10 & 36 $\pm$ 0 & 16 $\pm$ 0 & 22 $\pm$ 0 & \ul{15} $\pm$ 0 & 59 $\pm$ 0\\
\hline
\multirow{2}{*}{16} & min & 14 & 12 & \ul{9} & 27 & 25 & 15 & 19 & 16\\
& mean & 27 $\pm$ 9 & 31 $\pm$ 14 & 24 $\pm$ 11 & 27 $\pm$ 0 & 25 $\pm$ 0 & 15 $\pm$ 0 & 19 $\pm$ 0 & \ul{16} $\pm$ 0\\
\hline
\multirow{2}{*}{17} & min & 8 & \ul{4} & \ul{4} & 19 & 11 & 10 & 8 & 5\\
& mean & 13 $\pm$ 2 & 12 $\pm$ 5 & 11 $\pm$ 4 & 19 $\pm$ 0 & 11 $\pm$ 0 & 10 $\pm$ 0 & 8 $\pm$ 0 & \ul{5} $\pm$ 0\\
\hline
\multirow{2}{*}{18} & min & 10 & 8 & 9 & 22 & 30 & 16 & \ul{7} & 27\\
& mean & 25 $\pm$ 9 & 25 $\pm$ 11 & 23 $\pm$ 9 & 22 $\pm$ 0 & 30 $\pm$ 0 & 16 $\pm$ 0 & \ul{7} $\pm$ 0 & 27 $\pm$ 0\\
\hline
\multirow{2}{*}{19} & min & 6 & \ul{3} & \ul{3} & 5 & 9 & 10 & 12 & \ul{3}\\
& mean & 12 $\pm$ 5 & 14 $\pm$ 6 & 12 $\pm$ 7 & 5 $\pm$ 0 & 9 $\pm$ 0 & 10 $\pm$ 0 & 12 $\pm$ 0 & \ul{3} $\pm$ 0\\
\hline
\multirow{2}{*}{20} & min & 6 & 5 & 4 & 7 & 7 & 7 & 6 & \ul{2}\\
& mean & 8 $\pm$ 1 & 8 $\pm$ 2 & 7 $\pm$ 2 & 7 $\pm$ 0 & 7 $\pm$ 0 & 7 $\pm$ 0 & 6 $\pm$ 0 & \ul{2} $\pm$ 0\\
\hline
\multirow{2}{*}{21} & min & 9 & 11 & 11 & 24 & 20 & \ul{8} & 21 & 19\\
& mean & 20 $\pm$ 8 & 22 $\pm$ 8 & 20 $\pm$ 8 & 24 $\pm$ 0 & 20 $\pm$ 0 & \ul{8} $\pm$ 0 & 21 $\pm$ 0 & 19 $\pm$ 0\\
\hline
\multirow{2}{*}{22} & min & \ul{15} & 17 & 18 & 20 & 62 & 50 & 33 & 49\\
& mean & 41 $\pm$ 20 & 42 $\pm$ 18 & 40 $\pm$ 18 & \ul{20} $\pm$ 0 & 62 $\pm$ 0 & 50 $\pm$ 0 & 33 $\pm$ 0 & 49 $\pm$ 0\\
\hline
\multirow{2}{*}{23} & min & \ul{4} & \ul{4} & \ul{4} & 9 & 5 & 7 & 5 & 5\\
& mean & 7 $\pm$ 2 & 8 $\pm$ 3 & 7 $\pm$ 3 & 9 $\pm$ 0 & \ul{5} $\pm$ 0 & 7 $\pm$ 0 & \ul{5} $\pm$ 0 & \ul{5} $\pm$ 0\\
\hline
\multirow{2}{*}{24} & min & \ul{13} & \ul{13} & 15 & 73 & 33 & 21 & 28 & 32\\
& mean & 29 $\pm$ 10 & 31 $\pm$ 11 & 29 $\pm$ 10 & 73 $\pm$ 0 & 33 $\pm$ 0 & \ul{21} $\pm$ 0 & 28 $\pm$ 0 & 32 $\pm$ 0\\
\hline
\end{xtabular}
}
\end{center}

\newpage

\begin{center}
{
\tiny
\tablecaption{\label{tab_cpu_time} CPU time comparison of the initialization methods}
\begin{xtabular}{c|c|r|r|r|r|r|r|r|r}
\hline
 & & \texttt{F} & \texttt{M} & \texttt{K} & \texttt{X} & \texttt{V} & \texttt{P} & \texttt{OV} & \texttt{OP}\\
\hline
\hline
\multirow{2}{*}{1} & min & 20 & 10 & 10 & 40 & 20 & 20 & 20 & 10\\
& mean & 43 $\pm$ 7 & 33 $\pm$ 10 & 24 $\pm$ 9 & 40 $\pm$ 0 & 20 $\pm$ 0 & 20 $\pm$ 0 & 20 $\pm$ 0 & 10 $\pm$ 0\\
\hline
\multirow{2}{*}{2} & min & 0 & 0 & 0 & 0 & 0 & 0 & 0 & 0\\
& mean & 0 $\pm$ 1 & 0 $\pm$ 1 & 0 $\pm$ 2 & 0 $\pm$ 0 & 0 $\pm$ 0 & 0 $\pm$ 0 & 0 $\pm$ 0 & 0 $\pm$ 0\\
\hline
\multirow{2}{*}{3} & min & 0 & 0 & 0 & 0 & 0 & 0 & 0 & 0\\
& mean & 0 $\pm$ 1 & 0 $\pm$ 1 & 0 $\pm$ 0 & 0 $\pm$ 0 & 0 $\pm$ 0 & 0 $\pm$ 0 & 0 $\pm$ 0 & 0 $\pm$ 0\\
\hline
\multirow{2}{*}{4} & min & 0 & 0 & 0 & 0 & 0 & 0 & 0 & 0\\
& mean & 0 $\pm$ 2 & 0 $\pm$ 1 & 0 $\pm$ 2 & 0 $\pm$ 0 & 0 $\pm$ 0 & 0 $\pm$ 0 & 0 $\pm$ 0 & 0 $\pm$ 0\\
\hline
\multirow{2}{*}{5} & min & 0 & 0 & 0 & 0 & 0 & 0 & 0 & 0\\
& mean & 0 $\pm$ 1 & 0 $\pm$ 1 & 0 $\pm$ 1 & 0 $\pm$ 0 & 0 $\pm$ 0 & 0 $\pm$ 0 & 0 $\pm$ 0 & 0 $\pm$ 0\\
\hline
\multirow{2}{*}{6} & min & 0 & 0 & 0 & 0 & 0 & 0 & 0 & 0\\
& mean & 0 $\pm$ 1 & 0 $\pm$ 2 & 1 $\pm$ 2 & 0 $\pm$ 0 & 0 $\pm$ 0 & 0 $\pm$ 0 & 0 $\pm$ 0 & 0 $\pm$ 0\\
\hline
\multirow{2}{*}{7} & min & 0 & 0 & 0 & 0 & 0 & 0 & 0 & 0\\
& mean & 0 $\pm$ 2 & 0 $\pm$ 2 & 0 $\pm$ 2 & 0 $\pm$ 0 & 0 $\pm$ 0 & 0 $\pm$ 0 & 0 $\pm$ 0 & 0 $\pm$ 0\\
\hline
\multirow{2}{*}{8} & min & 0 & 0 & 0 & 0 & 0 & 0 & 0 & 0\\
& mean & 0 $\pm$ 0 & 0 $\pm$ 1 & 0 $\pm$ 1 & 0 $\pm$ 0 & 0 $\pm$ 0 & 0 $\pm$ 0 & 0 $\pm$ 0 & 0 $\pm$ 0\\
\hline
\multirow{2}{*}{9} & min & 1690 & 1630 & 1580 & 2570 & 6920 & 12160 & 5040 & 12460\\
& mean & 3691 $\pm$ 1229 & 3370 $\pm$ 1178 & 3397 $\pm$ 1055 & 2570 $\pm$ 0 & 6920 $\pm$ 0 & 12160 $\pm$ 0 & 5040 $\pm$ 0 & 12460 $\pm$ 0\\
\hline
\multirow{2}{*}{10} & min & 0 & 10 & 10 & 50 & 30 & 50 & 10 & 50\\
& mean & 30 $\pm$ 9 & 32 $\pm$ 11 & 32 $\pm$ 10 & 50 $\pm$ 0 & 30 $\pm$ 0 & 50 $\pm$ 0 & 10 $\pm$ 0 & 50 $\pm$ 0\\
\hline
\multirow{2}{*}{11} & min & 380 & 350 & 320 & 670 & 960 & 790 & 620 & 810\\
& mean & 710 $\pm$ 174 & 673 $\pm$ 168 & 724 $\pm$ 166 & 670 $\pm$ 0 & 960 $\pm$ 0 & 790 $\pm$ 0 & 620 $\pm$ 0 & 810 $\pm$ 0\\
\hline
\multirow{2}{*}{12} & min & 10 & 10 & 10 & 40 & 20 & 10 & 30 & 20\\
& mean & 19 $\pm$ 7 & 19 $\pm$ 7 & 21 $\pm$ 6 & 40 $\pm$ 0 & 20 $\pm$ 0 & 10 $\pm$ 0 & 30 $\pm$ 0 & 20 $\pm$ 0\\
\hline
\multirow{2}{*}{13} & min & 20 & 20 & 20 & 60 & 20 & 80 & 30 & 40\\
& mean & 52 $\pm$ 17 & 52 $\pm$ 17 & 57 $\pm$ 20 & 60 $\pm$ 0 & 20 $\pm$ 0 & 80 $\pm$ 0 & 30 $\pm$ 0 & 40 $\pm$ 0\\
\hline
\multirow{2}{*}{14} & min & 10 & 10 & 10 & 30 & 30 & 210 & 20 & 200\\
& mean & 22 $\pm$ 7 & 21 $\pm$ 5 & 25 $\pm$ 8 & 30 $\pm$ 0 & 30 $\pm$ 0 & 210 $\pm$ 0 & 20 $\pm$ 0 & 200 $\pm$ 0\\
\hline
\multirow{2}{*}{15} & min & 50 & 50 & 50 & 140 & 60 & 140 & 70 & 280\\
& mean & 122 $\pm$ 50 & 126 $\pm$ 53 & 124 $\pm$ 41 & 140 $\pm$ 0 & 60 $\pm$ 0 & 140 $\pm$ 0 & 70 $\pm$ 0 & 280 $\pm$ 0\\
\hline
\multirow{2}{*}{16} & min & 0 & 0 & 0 & 10 & 10 & 10 & 10 & 10\\
& mean & 9 $\pm$ 5 & 11 $\pm$ 6 & 8 $\pm$ 5 & 10 $\pm$ 0 & 10 $\pm$ 0 & 10 $\pm$ 0 & 10 $\pm$ 0 & 10 $\pm$ 0\\
\hline
\multirow{2}{*}{17} & min & 0 & 0 & 0 & 0 & 0 & 0 & 0 & 0\\
& mean & 0 $\pm$ 2 & 1 $\pm$ 2 & 1 $\pm$ 2 & 0 $\pm$ 0 & 0 $\pm$ 0 & 0 $\pm$ 0 & 0 $\pm$ 0 & 0 $\pm$ 0\\
\hline
\multirow{2}{*}{18} & min & 40 & 30 & 30 & 50 & 100 & 70 & 30 & 100\\
& mean & 87 $\pm$ 27 & 87 $\pm$ 37 & 79 $\pm$ 26 & 50 $\pm$ 0 & 100 $\pm$ 0 & 70 $\pm$ 0 & 30 $\pm$ 0 & 100 $\pm$ 0\\
\hline
\multirow{2}{*}{19} & min & 0 & 0 & 0 & 10 & 10 & 30 & 10 & 20\\
& mean & 11 $\pm$ 5 & 12 $\pm$ 6 & 11 $\pm$ 8 & 10 $\pm$ 0 & 10 $\pm$ 0 & 30 $\pm$ 0 & 10 $\pm$ 0 & 20 $\pm$ 0\\
\hline
\multirow{2}{*}{20} & min & 0 & 0 & 0 & 0 & 0 & 0 & 0 & 0\\
& mean & 0 $\pm$ 2 & 0 $\pm$ 2 & 0 $\pm$ 2 & 0 $\pm$ 0 & 0 $\pm$ 0 & 0 $\pm$ 0 & 0 $\pm$ 0 & 0 $\pm$ 0\\
\hline
\multirow{2}{*}{21} & min & 10 & 10 & 10 & 20 & 20 & 20 & 20 & 30\\
& mean & 21 $\pm$ 9 & 20 $\pm$ 8 & 20 $\pm$ 8 & 20 $\pm$ 0 & 20 $\pm$ 0 & 20 $\pm$ 0 & 20 $\pm$ 0 & 30 $\pm$ 0\\
\hline
\multirow{2}{*}{22} & min & 10 & 20 & 20 & 10 & 60 & 50 & 40 & 40\\
& mean & 40 $\pm$ 19 & 40 $\pm$ 18 & 38 $\pm$ 17 & 10 $\pm$ 0 & 60 $\pm$ 0 & 50 $\pm$ 0 & 40 $\pm$ 0 & 40 $\pm$ 0\\
\hline
\multirow{2}{*}{23} & min & 0 & 0 & 0 & 0 & 0 & 0 & 0 & 0\\
& mean & 0 $\pm$ 1 & 0 $\pm$ 1 & 0 $\pm$ 0 & 0 $\pm$ 0 & 0 $\pm$ 0 & 0 $\pm$ 0 & 0 $\pm$ 0 & 0 $\pm$ 0\\
\hline
\multirow{2}{*}{24} & min & 0 & 0 & 0 & 10 & 0 & 0 & 10 & 0\\
& mean & 6 $\pm$ 5 & 6 $\pm$ 5 & 6 $\pm$ 5 & 10 $\pm$ 0 & 0 $\pm$ 0 & 0 $\pm$ 0 & 10 $\pm$ 0 & 0 $\pm$ 0\\
\hline
\end{xtabular}
}
\end{center}

\newpage

\begin{table}
\centering
\caption{\label{tab_summary} Overall performance comparison of the initialization methods}
\begin{tabular}{c|c|r|r|r|r|r|r|r|r}
\hline
Statistic & Criterion & \texttt{F} & \texttt{M} & \texttt{K} & \texttt{X} & \texttt{V} & \texttt{P} & \texttt{OV} & \texttt{OP}\\
\hline
\hline
Min & Init.\ SSE & 2.968 & 1.418 & 1.348 & 2.184 & 1.107 & 1.043 & 1.067 & \ul{1.002}\\
 & Final SSE & 1.001 & 1.003 & \ul{1.000} & 1.163 & 1.019 & 1.018 & 1.031 & 1.005\\
 & \# Iters.\ & 1.488 & 1.284 & \ul{1.183} & 2.978 & 2.469 & 2.013 & 2.034 & 1.793\\
\hline
Mean & Init.\ SSE & 3.250 & 2.171 & 1.831 & 2.184 & 1.107 & 1.043 & 1.067 & \ul{1.002}\\
 & Final SSE & 1.047 & 1.049 & 1.032 & 1.161 & 1.017 & 1.016 & 1.029 & \ul{1.003}\\
 & \# Iters.\ & 2.466 & 2.528 & 2.314 & 2.581 & 2.057 & 1.715 & 1.740 & \ul{1.481}\\
\hline
Stdev & Init.\ SSE & \ul{1.000} & 13.392 & 12.093 & -- & -- & -- & -- & --\\
 & Final SSE & \ul{1.013} & 1.239 & 1.172 & -- & -- & -- & -- & --\\
 & \# Iters.\ & \ul{1.000} & 1.166 & 1.098 & -- & -- & -- & -- & --\\
\hline
\end{tabular}
\end{table}

\begin{table}[ht]
\linespread{1}
\centering
\normalsize
\caption{\label{tab_rel_var} Performance of Otsu Var-Part relative to Var-Part (\texttt{V})}
\begin{tabular}{@{\extracolsep{\fill}} crrr}
\hline
Criterion & Worse than \texttt{V} & Same as \texttt{V} & Better than \texttt{V}\\
\hline
\hline
Init.\ SSE & 6 & 3 & 15\\
Final SSE & 6 & 16 & 2\\
\# Iters.\ & 8 & 3 & 13\\
\hline
\end{tabular}
\end{table}

\begin{table}[ht]
\linespread{1}
\centering
\normalsize
\caption{\label{tab_rel_pca} Performance of Otsu PCA-Part relative to PCA-Part (\texttt{P})}
\begin{tabular}{@{\extracolsep{\fill}} crrr}
\hline
Criterion & Worse than \texttt{P} & Same as \texttt{P} & Better than \texttt{P}\\
\hline
\hline
Init.\ SSE & 1 & 2 & 21\\
Final SSE & 4 & 14 & 6\\
\# Iters.\ & 6 & 1 & 17\\
\hline
\end{tabular}
\end{table}

\end{document}